\documentclass[a4paper,conference]{IEEEtran}
\IEEEoverridecommandlockouts
\usepackage{cite}
\usepackage{amsmath,amssymb,amsfonts}
\pdfoutput=1
\usepackage{graphicx}
\usepackage{textcomp}
\usepackage{xcolor}
\def\BibTeX{{\rm B\kern-.05em{\sc i\kern-.025em b}\kern-.08em
    T\kern-.1667em\lower.7ex\hbox{E}\kern-.125emX}}
\usepackage{xspace}
\newcommand{\myAlg}{DCFL\xspace}
\usepackage{algorithm}
\usepackage{algpseudocode}
\usepackage{multirow}
\usepackage{graphicx}
\usepackage{textcomp}
\usepackage{epstopdf}
\usepackage{booktabs}
\usepackage{threeparttable}
\usepackage{subfigure}
\usepackage{float}

\makeatletter

\newcommand{\Rmnum}[1]{\expandafter\@slowromancap\romannumeral #1@}
\makeatother

\usepackage{algorithm}
\usepackage{algpseudocode}
\usepackage{indentfirst} 
\usepackage{xspace}

\begin{document}

\title{DCFL: Non-IID awareness Data Condensation aided Federated Learning}

\author{
    Shaohan Sha\textsuperscript{\rm *, \rm 1} ,
    Yafeng Sun\textsuperscript{\rm *, \rm 1} 
    \\
    \vspace{1ex}
    $^{1}$Jilin University \quad \\
    \thanks{$^{*}$Equal Contribution.}
    \thanks{}%
}

\maketitle

\begin{abstract}
Federated learning is a decentralized learning paradigm wherein a central server trains a global model iteratively by utilizing clients who possess a certain amount of private datasets.
The challenge lies in the fact that the client-side private data may not be identically and independently distributed, significantly impacting the accuracy of the global model.
Existing methods commonly address the Non-IID challenge by focusing on optimization, client selection and data complement.
However, most approaches tend to overlook the perspective of the private data itself due to privacy constraints.
Intuitively, statistical distinctions among private data on the client side can help mitigate the Non-IID degree.
Besides, the recent advancements in dataset condensation technology have inspired us to investigate its potential applicability in addressing Non-IID issues while maintaining privacy.
Motivated by this, we propose \myAlg which divides clients into groups by using the Centered Kernel Alignment (CKA) method, then uses dataset condensation methods with non-IID awareness to complete clients.
The private data from clients within the same group is complementary and their condensed data is accessible to all clients in the group.
Additionally, CKA-guided client selection strategy, filtering mechanisms, and data enhancement techniques are incorporated to efficiently and precisely utilize the condensed data, enhance model performance, and minimize communication time.
Experimental results demonstrate that \myAlg achieves competitive performance on popular federated learning benchmarks including MNIST, FashionMNIST, SVHN, and CIFAR-10 with existing FL protocol.
\end{abstract}


\section{Introduction}
\label{sec:introduction}

\begin{figure*}[!ht]
    \centering
    \includegraphics[width=2\columnwidth]{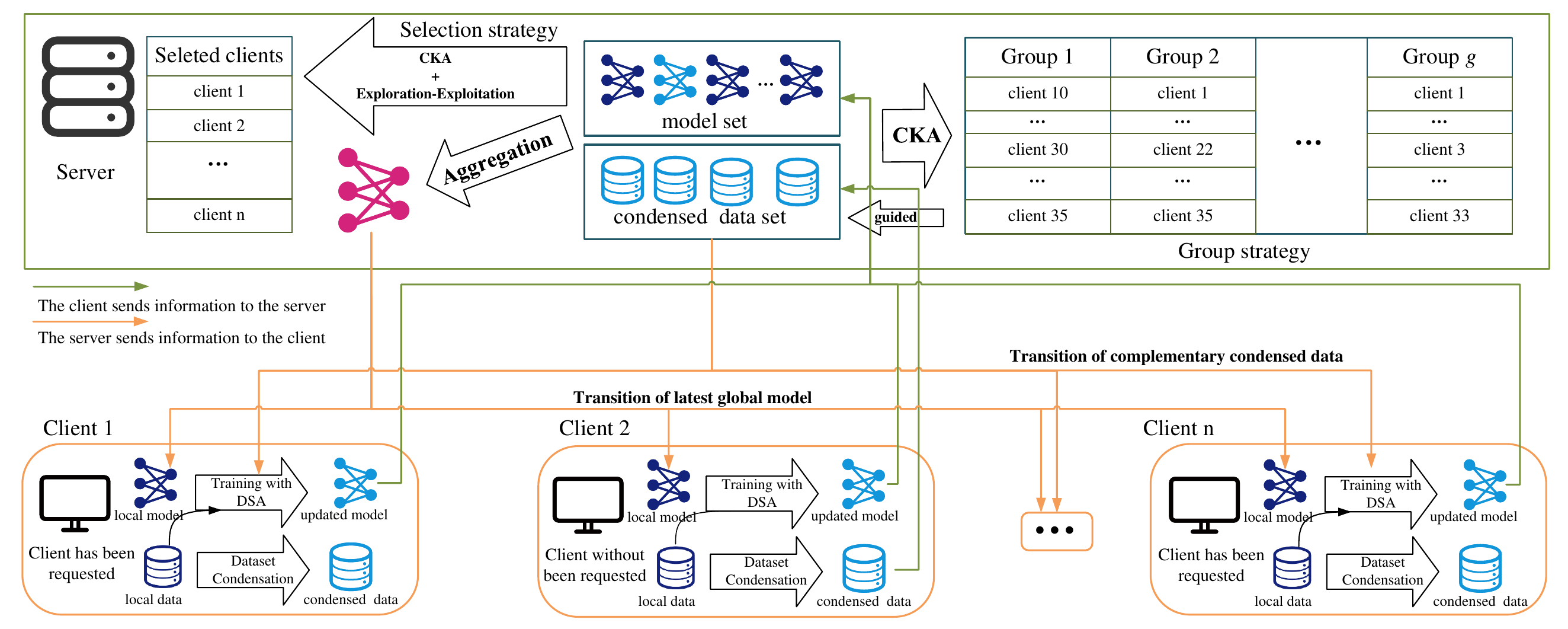}
    \caption{The framework of \myAlg}
    \label{fig::flowchart}
\end{figure*}

With the proliferation of Internet of Things  devices, Federated learning (FL) has emerged as a promising machine learning paradigm. In FL, many clients collaboratively train a model under the orchestration of a central server, while keeping the training data local~\cite{mcmahan2017communication}.
Due to FL’s great privacy protection, communication reduction for processing voluminous distributed data, high scalability, and other excellent advantages~\cite{wang2020convergence}, it has received extensive attention in academia and industry.
At present, FL has been widely used in Keyboard Word Spotting~\cite{li2022avoid}, Diver Activity Recognition (DAR)~\cite{doshi2022federated}, Speech Recognition~\cite{yang2019federated}, Health Monitoring~\cite{xu2021federated}, and other domains.

However, Non-Identical and Independently Distribution~\cite{kairouz2021advances} (note: aka statistical heterogeneous, data heterogeneity or Non-IID), as the distribution of the client’s local data is not representative of the distribution of the overall data~\cite{li2022federated}, results in FL tasks to become more complex and meet some challenges.
The Non-IID issue causes the training process of the server model to more fluctuate, leads to more communication rounds between clients and server, and degrades the final model performance drastically~\cite{hsieh2020non}.
According to Ref.~\cite{zhao2018federated}, the reduction in the test accuracy of FedAvg for Non-IID data up to 11.31\% in MNIST~\cite{lecun1998mnist}, 51.31\% in CIFAR-10~\cite{krizhevsky2009learning}, 54.5\% in KWS~\cite{warden2018speech}.

Recently, researchers have devised FL methods from diverse perspectives to mitigate the negative effects caused by data heterogeneity:

From the perspective of the FL optimization method, Li et al.~\cite{li2020federated} proposed FedProx, which is based on FedAvg and introduces an additional L2 regularization penalty term in the local objective function to restrict the disturbance to the global server model aggregation from participated clients whose model trained on the highly Non-IID local dataset.
Wang et al.~\cite{wang2020tackling} proposed FedNova, which adopts a normalized averaging method that eliminated objective inconsistency.
Karimireddy et al.~\cite{karimireddy2020scaffold} proposed Scaffload which uses control variates or variance reduction to correct for the client-drift caused by data heterogeneous in its local updates.
While those methods outperform FedAvg and contemporary FL methods in Non-IID scenarios, however, these methods have limitations: They don’t consistently outperform other algorithms in all Non-IID data settings~\cite{li2022federated}, and they can’t influence the inherent Non-IID properties of clients. Besides, they adopt the traditional random participant selection strategy to select clients without considering the data distribution complementary of clients.

From the perspective of client selection strategies, there are many works that mainly focus on System Heterogeneity, such as computation capacity, communication capacity, or both, in order to get as many clients as possible involved in the training process and reduce communication time~\cite{soltani2022survey}.
Few of them try to select clients based on analyzing the statistically complementary relationship between clients, like the work of~\cite{ma2021client} who divides clients into several groups based on Group Earth Mover’s Distance(GEMD), the work of~\cite{zhao2018federated} utilizes weight divergence to recognize the data heterogeneity degree of client data.
To some extent, those works take advantage of the heterogeneous information of the client's private data, but they don’t solve the inherent Non-IID issue.
From the perspective of data completion, Zhao Y et al.~\cite{zhao2018federated} propose only 5\% globally shared data can increase the accuracy of the CIFAR-10 dataset up to 30\%, Ma J et al.~\cite{ma2021client} proposes to share a distribution information version of client private local dataset which alleviate the negative effect of Non-IID data on the training performance in some degree.
Data complementation works directly from the data itself to achieve better performance.
However, the above approaches compromise the fundamental principles of user privacy protection and are thus limited to specific FL scenarios.
In most practical settings, accessing client data or local data distribution information as supplementary data is prohibited.

In conclusion, while data complementarity is a promising solution to alleviate the Non-IID problem~\cite{zhao2018federated}, its communication burden and privacy challenges still hinder its widespread application.
To this end, we propose a novel execution framework of federated learning, called $\myAlg$, which stands for Data Condensation aided Federated Learning with Non-IID awareness.
\myAlg aims to mitigate the negative impacts of Non-IID data on FL model training, communication, and performance by using condensed data efficiently and precisely.
Instead of the conventional scheme where the server randomly selects a fraction of clients to participate in FL training ~\cite{mcmahan2017communication}~\cite{li2020federated}~\cite{wang2020tackling}~\cite{karimireddy2020scaffold}, we take a novel viewpoint on client data complementarity and use that information to guide participant selection.
Measuring the complementarity of client data is a crucial challenge since accessing or transferring client data distribution is prohibited due to privacy and confidentiality concerns~\cite{kairouz2021advances}.
Centered Kernel Alignment (CKA) is a method that can be used to measure the similarity between representational layers of different clients' local models~\cite{kornblith2019similarity}, making it an ideal fit for assessing the complementarity of client data.
Specifically, DCFL starts by maintaining a table that records the correlation between each client model and other selected client models.
Further, we endeavor to utilize the table to facilitate the training of the local model by selected clients with relatively less bias and full knowledge, through the utilization of auxiliary data from other clients. The deeper problem is how to get the auxiliary data from complemented clients.
In the traditional FL implementation process, we can only get weights or gradients from participating clients.
Motivated by condensed data obtained by data condensation methods owns informative, representative, small quantity features and no fear of privacy violation, we resort to utilizing condensate data from selected clients as a backbone to be delivered between the server and clients.
Besides, Data filtering and data augmentation are also employed, to utilize the condensed data with Non-IID awareness, further improve the final model performance, and reduce the total communication rounds in the implementation of \myAlg.

Our contributions in this work are summarized hereafter:

\begin{itemize}
\item CKA-based client complementarity.
The CKA method is introduced to obtain the complementarity between clients, which guides client selection and condensed data transfer in \myAlg.
Specifically, the server-side calculates complementarity between each client and all other clients utilizing CKA, then the clients are grouped according to the complementarity.
According to that, we can fine-grained select participating clients, further reduce the overall communication cost, and achieve better final model performance.
To the best of our knowledge, the proposed \myAlg is the first effort to apply CKA to client grouping, client selection, and condensed data utilization.

\item Condensed data-assisted Client model training with Non-IID awareness.
When the client model is training, real data from the client and the complement condensed data with or without filtering by the server from the same complement group’s clients collaborate.
Additionally, the DSA data augmentation technique, which is popular in data condensation methods, is used throughout the training process and the participating clients’ weight calculation formula on the server side is reorganized according to changes in clients’ local dataset quantity.
The above mechanisms not only further reduce the communication wheel, making the training process more stable but also ultimately achieve a better final model performance.

\item We employed four public datasets MNIST, Fashion MNIST, SVHN, and CIFAR-10, to validate the effectiveness of the \myAlg algorithm proposed in this paper.
Our code is built upon the lightweight and highly customizable framework - FedLab~\cite{zeng2023fedlab}.
It's worth noting that our code is publicly available and can be easily reproduced.
\end{itemize}

The rest of the paper is organized as follows.
Section \Rmnum{2} introduces the overview of DCFL.
We present the design detail and theory proofs in Section \Rmnum{3}.
Section \Rmnum{4} presents the simulation results under different scenarios.
In Section \Rmnum{5}, we describe the related work about DCFL.
Finally we conclude the paper in Section \Rmnum{6}.

\section{Design details}
In this part, we present the design details of \myAlg in four aspects:
\begin{table*}[t]

\caption{\centering{Notation descriptions.}}
\label{table::notation_table}

\vskip 0.15in
\begin{center}
\begin{small}
\begin{tabular}{l|l}
\toprule
Notation & Description \\ \midrule
$K$ & The total client number in FL system \\
$D_k$ & Local training set of client $k$\\
$\widetilde{\mathcal{D}}_k$ & Condensed data set of client $k$\\
$\widetilde{\mathcal{D}}$ & Condensed data set which received by the server\\
$\eta_{c}$ & The learning rate when using clients' local data\\
$\eta_{s}$ & The learning rate when using clients' condensed data\\
$B_{c}$ & Batch size for client's local private data\\
$B_{s}$ & Batch size for client's obtained condensed data\\
$E_{c}$ & The number of local epochs by using local private data\\
$E_{s}$ & The number of local epochs by using obtained condensed data\\
$w^{t}_k$ & Local model of client $k$ at round $t$\\
$w^{c}_{k}$ & The classifier of client $k$'s model \\
w$^{t}$ & Global model at round $t$ \\
$\mathcal{A}_{w}$ & Differentiable augmentation which parameterized with $w$\\
$C_{com}$ & The fraction of clients who were selected to take part in the FL training process \\
$C_{pre}$ & The fraction of clients who were selected to pre-train server model and provide information \\
$M$ & The total requests rounds for server model pre-training \\
$T$ & The total communication rounds between server and clients \\
$\mathcal{M}$ & The CKA valuation matrix between clients \\
$\mathbb{C}_{opt}$ & Clients who are selected to take part in process\\
$\mathbb{C}_{E}$ & Clients who have been selected to take part in process\\
$\epsilon$ & Exploitation factor(between 0 and 1)\\
\bottomrule
\end{tabular}
\end{small}
\end{center}
\vskip -0.2in
\end{table*}

 
 
\begin{algorithm}
    \caption {The Server execution flow of DCFL.} 
    \label{alg::algorithm_server}
    \begin{algorithmic}[1] 
        \State initialize global model w$^{0}$; $\mathbb{C}_{E}\leftarrow\emptyset$
        \State // the pre-training stage of server model
        \For{$r = 1$ to $M$}
            \State $n$ $\leftarrow$ max($C_{pre}$ * $K$, 1) 
            \State $\mathbb{C}_{opt}$ $\leftarrow$ (random select $n$ clients from $K$ clients)
            \For{$k$ $\, \in \, $ $\mathbb{C}_{opt}$ $\mathbf{in \  parallel}$} 
                \If{$k$ not in $\mathbb{C}_{E}$}
                    \State  $w_{k}^{c}$, $\widetilde{\mathcal{D}}_{k}$ $\leftarrow$ ClientUpdateWoCD($k$, w$^{0}$)
                    \State  $\mathbb{C}_{E}$ $\leftarrow$ UpdateClient($\mathbb{C}_{E}$, $k$)
                \EndIf
            \EndFor
            \State $\widetilde{\mathcal{D}}$ $\leftarrow$ aggregate($\widetilde{\mathcal{D}}_{1}$, $\widetilde{\mathcal{D}}_{2}$, $\ldots$, $\widetilde{\mathcal{D}}_{n}$)
            \State $\mathcal{M}$ $\leftarrow$ updateCKAMatrix($w_{1}^{c}$, $w_{2}^{c}$, $\ldots$, $w_{n}^{c}$)
        \EndFor
        \State w$^{1}$ $\Leftarrow$ ServerUpdate(w$^{0}$, $\widetilde{\mathcal{D}}$)
        \State // the training stage of server model
        \For{$r = 1$ to $T$} 
            \State $n$ $\leftarrow$ max($C_{com}$ * $K$, 1) 
            \State  $\mathbb{C}^{*}$ = SelectForExploit($\mathbb{C}_{E}$, $\epsilon * n$, $\mathcal{M}$)
            \State  $\mathbb{C}_{opt}$ = $\mathbb{C}^{*}$ $\cup$ SelectForExplore($\urcorner{\mathbb{C}_{E}}$, $(1 - \epsilon) * n$)
            \For{$k$ $\, \in \, $ $\mathbb{C}_{opt}$ $\mathbf{in \  parallel}$} 
                \If{$k$ not in $\mathbb{C}_{E}$}
                    \State  $w_{k}^{r}$, $\widetilde{\mathcal{D}}_{k}$ $\leftarrow$ ClientUpdateWoCD($k$, w$^{r}$)
                    \State$\mathbb{C}_{E}$ $\leftarrow$ UpdateClient($\mathbb{C}_{E}$, $k$)
                    \State$\widetilde{\mathcal{D}}$ $\leftarrow$ UpdateCondensedData($\widetilde{\mathcal{D}}_{k}$, $\widetilde{\mathcal{D}}$)
                \Else 
                    \State $\widetilde{\mathcal{D}}_{k}$ $\leftarrow$ GetComplementaryCondensedDataForK($k$, $\widetilde{\mathcal{D}}$, $\mathbb{C}^{E}$, $\mathcal{M}$)
                    \State $\widetilde{\mathcal{D}}_{k}^{'}$ $\leftarrow$ filterCondensedData($\widetilde{\mathcal{D}}_{k}$)
                    \State $w_{k}^{r}$ $\leftarrow$ ClientUpdateWCD($k$, w$^{r}$, $\widetilde{\mathcal{D}}_{k}^{'}$)
                \EndIf
            \EndFor
            \State p = getOptimizedWeights($\mathbb{C}_{opt}$)
            \State Server aggregates local model w$^{r+1}$ =  $\Sigma_{k=1}^{n} p_{k}w_{k}^{r}$
        \EndFor
        \State \Return w$^{T}$
    \end{algorithmic}
\end{algorithm}

\begin{algorithm}
   
    \caption {The Client execution flow of DCFL.} 
    \label{alg::algorithm_client}
    \begin{algorithmic}[1] 
        \Function{ClientUpdateWoCD}{$k$, w}
           \State initialize client $k$'s model with w: $w_{k}\leftarrow$ w
           \State $\mathcal{B} \leftarrow$ (split $D_{k}$ into batches of size $B_{c}$)
           \State $\widetilde{\mathcal{D}}_{k}$ $\leftarrow$ ClientDatasetDistillation(k, w, $D_{k}$)
            \For{each local epoch $i$ from 1 to $E_{c}$}\par
                \For {batch $b$ $\epsilon$ $\mathcal{B}$}
                    \State $w_{k}$ $\leftarrow$ $\eta_{c}$ $\triangledown$$\mathcal{L}(w_{k}; \mathcal{A}_w(b))$
                \EndFor
            \EndFor
            \State\Return $w_{k}$, $\widetilde{\mathcal{D}}_{k}$
        \EndFunction
        \Function{ClientUpdateWCD}{$k$, w, $\widetilde{\mathcal{D}}$}
            \State $\widetilde{\mathcal{D}}_{k}$ $\leftarrow$ UpdateCondensedData($\widetilde{\mathcal{D}}$, $\widetilde{\mathcal{D}}_{k}$)
            \State $w_{k} \leftarrow$ ClientUpdateWoCD($k$, w)
            
            \State //The fine-tuning stage
            \State $\mathcal{B} \leftarrow$ (split $\widetilde{\mathcal{D}}_{k}$ into batches of size $B_{s}$)
            \For{each local epoch $i$ from 1 to $E_{s}$}\par
                \For {batch $b$ $\epsilon$ $\mathcal{B}$}
                    \State $w_{k}$ $\leftarrow$ $\eta_{s}$ $\triangledown$$\mathcal{L}(w_{k}; \mathcal{A}_w(b))$
                \EndFor
            \EndFor
            \State\Return $w_{k}$
        \EndFunction
    \end{algorithmic}
\end{algorithm}
\subsection{CKA-based client complementarity}
Why we can use CKA to judge the data complementarity of clients? 
the original idea comes from weight divergence. 
In the work of~\cite{li2022federated} who uses 
\begin{equation}
\text {weight divergence}=\frac{\lVert w^{\text {FedAvg }}-w^{S G D} \rVert}{\lVert w^{S G D} \rVert}
\label{eq::eq_1}
\end{equation}
to calculate the weight divergence between FedAvg and SGD \footnote{SGD is the ideal situation where the server knows the overall dataset distribution and trains the whole dataset collected from all clients}, and finds there is an association between the weight divergence and the skewness of the data. 
While they heuristic demonstrate the root cause of the weight divergence is due to the distance between the data distribution on each client and the population distribution and considers partial and whole into account, they don’t further infer the data distribution relationship between peer-to-peer that clients with similar data distribution will have minimal weight divergence according to the variation of \eqref{eq::eq_1}, like 
\begin{equation}
    \text { weight divergence }=\frac{\lVert w_{t}^{m}-w_{t}^{n}\rVert} {\lVert w^{S G D}\rVert}
    \label{eq::eq_2}
\end{equation}
which using the weight of client m and client n in round t instead of FedAvg and SGD to perform calculation.
Then clients with different data distributions will have large weight divergence which can further infer their complementary relationship.
To further reflect weight divergence, Ref.~\cite{zhao2018federated}~\cite{ma2021client} propose applying the earth mover’s distance(EMD) as follows:

\begin{equation}
\sum_{i=1}^C\lVert p^{(k)}(y=i)-p(y=i)\rVert
\label{eq::eq_3}
\end{equation}
between the global and local data distribution (p and pk) to simulation.
The method is impractical because it assumes we already know the distribution of overall data distribution about different classes ($p(y=i)$) and we can’t obtain the specific data distribution about clients ($p^{k}(y=i)$) in consideration of privacy protection.
How to reflect weight divergence efficiently and reasonably, the work of~\cite{luo2021no} sheds light on this problem, applying CKA as the alternative measure method of weight divergence to expose how the data heterogeneity affects each layer of a deep classification model and find there exists a greater bias in the classifier than other layers.
While they further propose a novel algorithm Classifier Calibration with Virtual Representations(CCVR) and achieve excellent performance on CIFAR-10, CIFAR-100, and CINIC-10 datasets, they neglect to further infer the distribution relationship between clients and don’t further consider the use of CKA to guide client selection instead of traditional random client selection.
Motivated by the above discovery, we dig deeper to propose using CKA to measure weight divergence and data complementarity between clients which only uses the partial model parameters of the classifier to calculate so that further reduce communication bandwidth and computation time cost~\cite{ma2021client}.

\begin{figure}
    \centering
    \includegraphics[width=1\columnwidth]{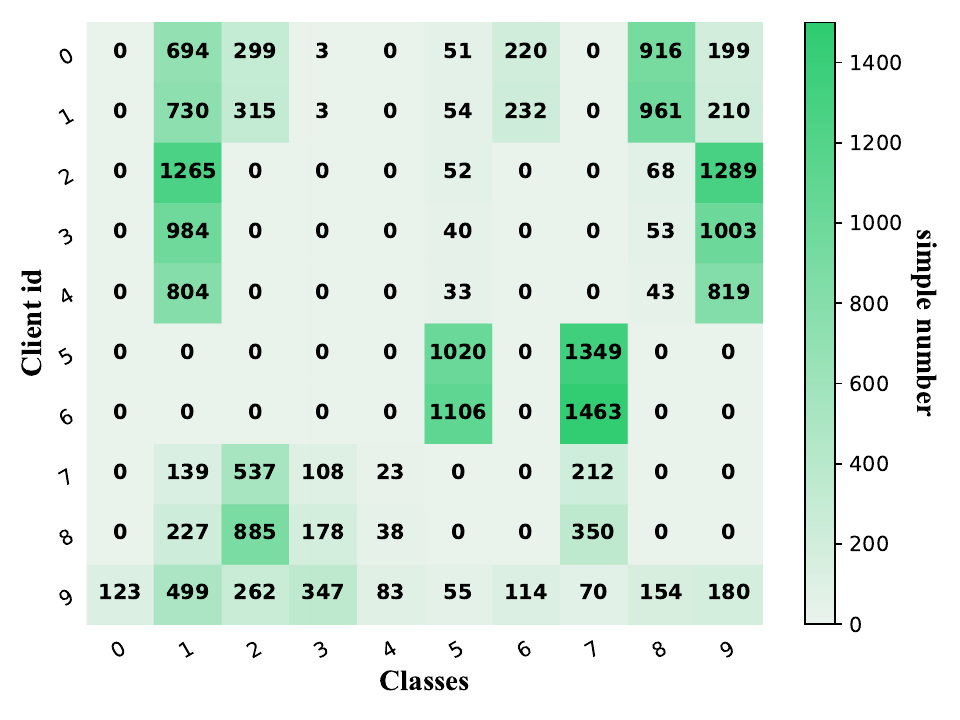}
    \caption{\centering{Label distribution}}
    \label{fig::label_distribution}
\end{figure}

To vividly understand and verify our assumption: the data similarity and complementary relationship between different clients can be derived from model weight similarity measurement methods, like CKA.
We perform an experimental study on clients with heterogeneous datasets.
For the sake of simplicity and representativeness, we chose CIFAR-10 with 10 clients and chose a convolutional neural network with four layers. 
As for the Non-IID setting, we partition the data according to the Dirichlet distribution, we set the 10 clients into 5 groups, and the group list which each group contains a certain amount of clients is [2, 3, 2, 2, 1]. The detailed data distribution of clients is shown in Figure \ref{fig::label_distribution}. 

\begin{figure}
    \centering
    \includegraphics[width=\columnwidth]{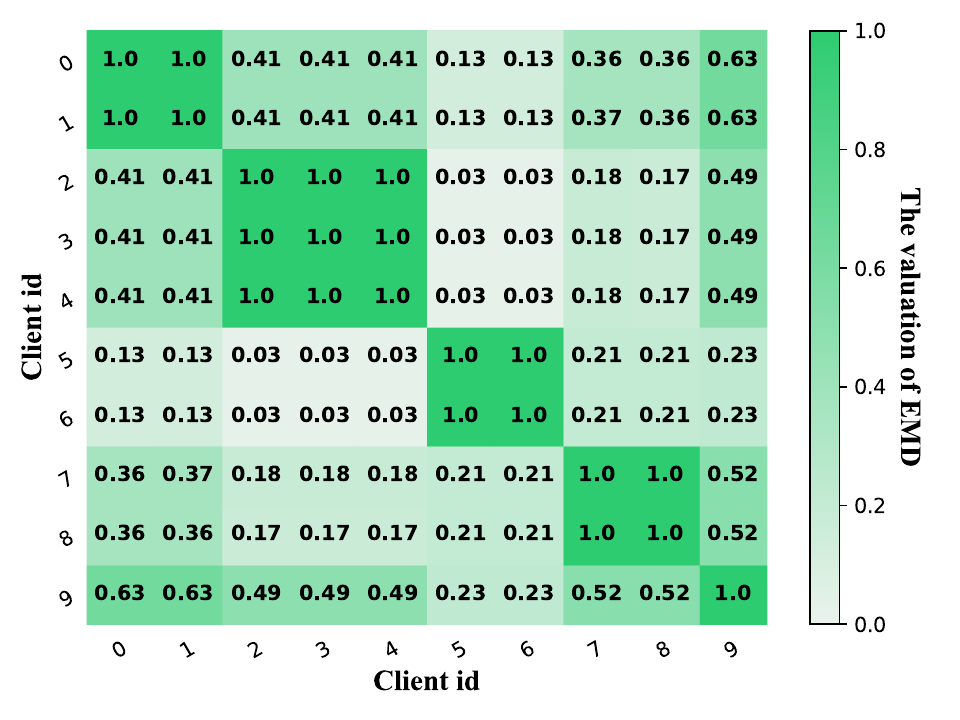}
    \caption{\centering{The EMD valuation between clients}}
    \label{fig::EMD_valuation}
\end{figure}

We first calculate the pairwise EMD valuations of those clients to reflect their data complementarity and show in Figure \ref{fig::EMD_valuation}, which is the optimal situation where the data distribution of each client is known.
From Figure \ref{fig::EMD_valuation}, we can find out that client belonging to the same group have symmetrical and relatively high EMD valuations, while clients with different data distributions will have relatively low valuations.
The lower valuation means more differences between clients’ data distribution. For example, we can see client 0 is similar to client 1 and is highly different from client 5 and client 6.

\begin{figure}
    \centering
    \includegraphics[width=\columnwidth]{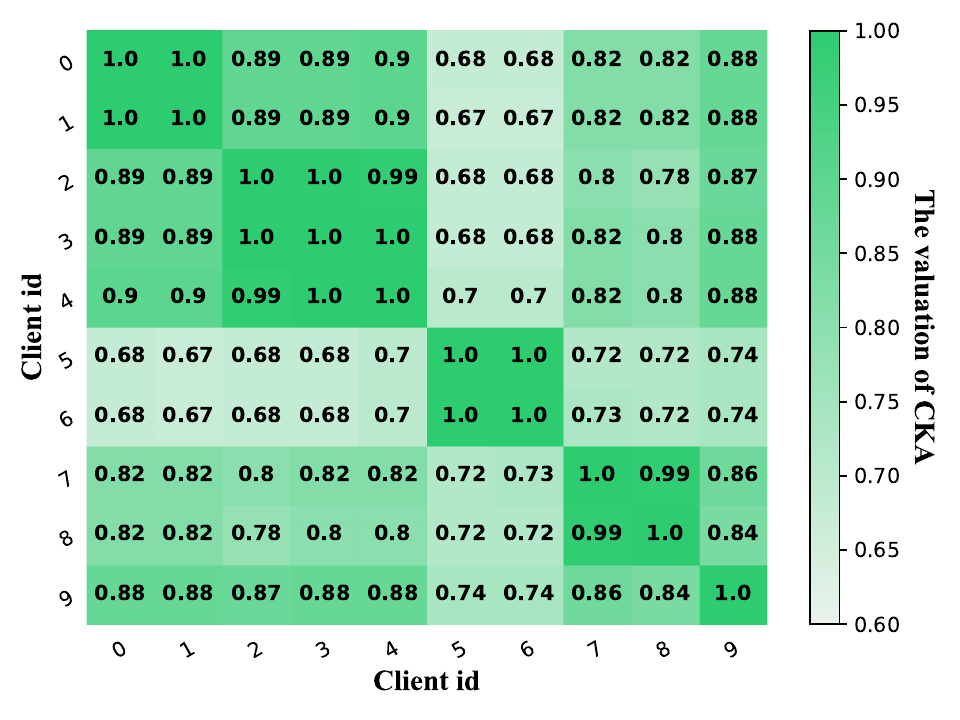}
    \caption{\centering{The CKA valuation between clients}}
    \label{fig::CKA_valuation}
\end{figure}

Then based on clients’ partial model parameters of classifier who have been updated for 10 epochs by using their local dataset to pair-wise calculate the CKA valuation, we can find the CKA relationship between clients in Figure \ref{fig::CKA_valuation}, who have similar judgements about clients’ data complementarity relationship like Figure \ref{fig::EMD_valuation}.
So our proposition that use CKA to measure the data complementarity relationship between clients is credible and pragmatic.

\subsection{CKA-guided client selection}
Instead of random client selection in the traditional Fl system, according to \eqref{eq::eq_4},we utilize the obtained CKA valuation matrix as the guide standard in \myAlg, in which clients with the highest CKA valuation sum value have a higher probability of being selected in the training process (Line 20 of Algorithm \ref{alg::algorithm_server}).
Because the client has a high sum CKA valuation the client’s private data distribution is more similar to other clients.

\begin{equation}
prob(C_{i}) = \frac{\sum_{\substack{j=1,\\j \neq i}}^{|{C}_{E}|} \mathcal{M}_{i \ j}}{\sum_{i=1}^{|{C}_{E}|}\sum_{\substack{j=1,\\j \neq i}}^{|{C}_{E}|} \mathcal{M}_{i \ j}}
\label{eq::eq_4}
\end{equation}

Besides, we also incorporate the exploration-exploitation mechanism in \myAlg(Line 20 – 21 of Algorithm \ref{alg::algorithm_server}), like the work of~\cite{li2022pyramidfl}, to further utilize the information obtained by the calculation of CKA between different clients and ensure the robustness of the server model.
The exploration ratio e approaches 1, the server tends to traditional random selection which selects unseen clients to take part in the FL training process, when the exploration ratio approaches 0, the server tends to use known knowledge to guide client selection.

\subsection{Transportation of condensed data with Non-IID awareness}
In \myAlg, we use condensed data, obtained from clients who implement data condensation in their private dataset, 
as auxiliary data to transport in two-phase: in the stage of server model pre-training (Line 4 – 14 of Algorithm \ref{alg::algorithm_server}). 
The server sends requests to clients, and then the clients can choose to send back information that contains condensed data or not according to their availability.
When the server obtains a certain quantity of information after T request rounds, 
it can then use those condensed data to train the model instead of traditional random weight model initialization to further reduce communication rounds; 
In the stage of client-server communication, selected clients who never take part in training process before will be asked to send condensate data and whole model parameter to the server for extending server’s knowledge (Line 23 – 26 of Algorithm \ref{alg::algorithm_server}). 
According to the complementary relationship, the server sends the latest model and condensed data from complementary clients to selected clients who have taken part in the training process before (Line 28 – 30 of Algorithm \ref{alg::algorithm_server}). 
Instead of transporting all collected condensed datasets, our method is more communication efficient and more fine-grained.

\subsection{Utilization of condensate data and data augmentation}
\myAlg is designed to operate condensate data with or without filtered 
(filter or not according to the quality of condensed data) 
for fine-tuning and applying the augmentation methods derived from DSA on the training process of the client side. We have introduced a filtration mechanism before the server delivers condensate data, 
which can retain more valuable data, further reduce communication bandwidth, and make the training process more stable.
If the filter ratio $r$ is not set equal to 0, 
condensate data can be filtered based on their performance by the server (Line 29 of Algorithm \ref{alg::algorithm_server}).
After participating clients receive the last server model and condensed data, they use their local dataset to update the model for $E_{c}$ epochs as traditionally and then use obtained condensed data from the server with a smaller learning rate $\eta_{s}$ to fine-tune for $E_{s}$ epochs (Line 16 – 22 of Algorithm \ref{alg::algorithm_client}).
Before we adopted this organization method, we tried different methods for using condensed data like mingling synthetic data and local private data as a whole for training or freezing model parameters and using synthetic data or real local data to fine-tune the classifier or more layers.
After comparing the aforementioned methods, we found the currently adopted training method is the most effective and beneficial for model training. 
We also find Differentiable Siamese Augmentation method proposed by DSA can also be applied in \myAlg to preprocess clients’ local training data and received synthetic data before training (Line 8 and Line 20 of Algorithm \ref{alg::algorithm_client}) which significantly improves the final model performance.

\section{experiments}

\renewcommand{\arraystretch}{1.5} 
\begin{table*}[tp]  
	\centering  


	\fontsize{15}{20}\selectfont    
        \caption{\centering{Test accuracy of FL methods with different level of non-iid partitioning}}
        \label{tab::performance_comparison}
        \resizebox{\textwidth}{!}{
	   \begin{threeparttable}

            \begin{tabular}{*{14}c}  
                \toprule
      \multirow{2}{*}{ }&
                \multirow{2}{*}{\textbf{Method}}&  
                \multicolumn{4}{c}{$\alpha = 0.5$}&
                    \multicolumn{4}{c}{$\alpha = 0.1$}&
                    \multicolumn{4}{c}{$C_{k}=2$ (pathological Non-IID)}                                 \\    
                    \cmidrule(lr){3-6} \cmidrule(lr){7-10} \cmidrule(lr){11-14}
                                   &                                  & MNIST        & Fashion MNIST & SVHN         & CIFAR-10     & MNIST        & Fashion MNIST & SVHN         & CIFAR-10     & MNIST        & Fashion MNIST & SVHN         & CIFAR-10     \\  \cmidrule(lr){1-2} \cmidrule(lr){3-6} \cmidrule(lr){7-10} \cmidrule(lr){11-14}
\multirow{2}{*}{\textbf{FedAvg}}   & Tradition                        & 97.39±0.18\%          & 87.19±0.33\%          & 86.17±0.33\%          & 49.15±0.94\%          & 95.03±0.72\%          & 77.77±1.67\%          & 69.87±2.09\%          & 30.24±1.50\%          & 71.79±3.82\%          & 54.28±3.40\%          & 74.03±2.30\%          & 35.86±2.27\% \\
                                   & DCFL                             & \textbf{98.47±0.07\%} & 88.68±0.18\%          & \textbf{92.19±0.20\%} & \textbf{63.88±0.41\%} & \textbf{97.82±0.11\%} & 84.44±0.28\%          & \textbf{87.72±0.45\%} & 46.63±0.46\%          & 92.88±0.80\%          & 69.14±2.39\%          & 85.01±0.62\%          & 52.73±0.60\%  \\ \cmidrule(lr){1-2} \cmidrule(lr){3-6} \cmidrule(lr){7-10} \cmidrule(lr){11-14}
                \morecmidrules\cmidrule(lr){1-2} \cmidrule(lr){3-6} \cmidrule(lr){7-10} \cmidrule(lr){11-14}
\multirow{2}{*}{\textbf{FedProx}}  & Tradition                        & 97.35±0.17\%          & 87.58±0.31\%          & 87.24±0.38\%          & 49.66±0.86\%          & 95.19±0.50\%          & 80.26±1.27\%          & 80.25±1.25\%          & 35.23±1.07\%          & 75.81±3.93\%          & 58.68±3.43\%          & 78.28±1.74\%          & 38.69±1.93\% \\
                                   & DCFL                             & 98.38±0.06\%          & 88.70±0.12\%          & 91.76±0.17\%          & 63.33±0.39\%          & 97.58±0.13\%          & 84.40±0.24\%          & 87.33±0.37\%          & \textbf{47.92±0.33\%} & \textbf{94.07±0.85\%} & \textbf{72.58±2.04\%} & \textbf{85.54±0.53\%} & \textbf{53.21±0.54\%} \\ \cmidrule(lr){1-2} \cmidrule(lr){3-6} \cmidrule(lr){7-10} \cmidrule(lr){11-14}
                \morecmidrules\cmidrule(lr){1-2} \cmidrule(lr){3-6} \cmidrule(lr){7-10} \cmidrule(lr){11-14}
\multirow{2}{*}{\textbf{FedNova}}  & Tradition                        & 97.37±0.16\%          & 87.21±0.34\%          & 86.28±0.30\%          & 49.30±1.04\%          & 95.68±0.42\%          & 78.78±1.53\%          & 63.25±2.93\%          & 27.57±1.96\%          & 74.16±5.56\%          & 54.67±4.52\%          & 72.46±2.74\%          & 35.67±2.04\% \\
                                   & DCFL                             & 98.42±0.07\%          & \textbf{88.85±0.29\%} & 91.72±0.14\%          & 62.86±0.42\%          & 97.81±0.14\%          & 84.05±0.26\%          & 85.62±0.75\%          & 46.85±0.44\%          & 92.31±1.76\%          & 70.66±2.15\%          & 84.66±1.19\%          & 52.27±0.47\%  \\  \cmidrule(lr){1-2} \cmidrule(lr){3-6} \cmidrule(lr){7-10} \cmidrule(lr){11-14}
                \morecmidrules\cmidrule(lr){1-2} \cmidrule(lr){3-6} \cmidrule(lr){7-10} \cmidrule(lr){11-14}
\multirow{2}{*}{\textbf{FedDisco}} & Tradition                        & 97.44±0.13\%          & 87.28±0.33\%          & 86.62±0.38\%          & 49.86±1.12\%          & 95.25±0.62\%          & 77.49±1.74\%          & 73.47±1.97\%          & 30.69±1.60\%          & 76.09±6.85\%          & 52.36±4.14\%          & 74.05±2.55\%          & 34.78±1.80\% \\
                                   & DCFL                             & 98.45±0.06\%          & 88.84±0.10\%          & 92.11±0.12\%          & 63.10±0.36\%          & 97.78±0.08\%          & \textbf{84.73±0.28\%} & 87.61±0.37\%          & 47.46±0.32\%          & 93.04±1.36\%          & 71.01±2.15\%          & 85.44±1.48\%          & 53.12±0.69\%
                                   
                                   \\ 
                                    \bottomrule
        \end{tabular}  
                
    	\end{threeparttable} 
        }
\end{table*}  
\renewcommand{\arraystretch}{1.5} 
\begin{table*}[tp]  
	\centering  
	\fontsize{6}{8}\selectfont
        \caption{\centering{Number of communication rounds to reach a target accuracy for DCFL and other FL optimization methods on SVHN dataset}}
        \label{tab::communication_comparison} 
        \resizebox{\textwidth}{!}{
        \begin{threeparttable}  
            
            \begin{tabular}{c|c|ccc ccc ccc}  
                \hline
       
                    \multirow{2}{*}{ }&
                    \multirow{2}{*}{\textbf{Method$\backslash$ToA@}}&  
                    \multicolumn{3}{c}{$\alpha = 0.5$}&
                    \multicolumn{3}{c}{$\alpha = 0.1$}&
                    \multicolumn{3}{c}{$C_{k}=2$ (pathological Non-IID)} \cr  
                  \cline{3-5} \cline{6-8} \cline{9-11}
                    
                &&ToA@0.86&ToA@0.87&Accuracy&ToA@0.66&ToA@0.80&Accuracy &ToA@0.73&ToA@0.78&Accuracy
                    \cr  
                       \cline{1-2} \cline{3-5} \cline{6-8} \cline{9-11}
                    \multirow{4}{*}{\textbf{Tradition}}
                    &FedAvg& 75 & -  & 86.20\%  & 57 & -  & 70.32\% & 72 & -  & 76.19\% \cr
       
                &FedProx  &21 & 62 & 87.38 \% & 11 & 91 & 80.24\% & 35 & 72 & 78.76 \%\cr
       
                &FedNova &91 & -  & 86.27 \% & 85 & -  & 66.53\% & 92 & -  & 73.54\% \cr  
       
                &FedDisco & 43 & -  & 86.49 \% & 31 & -  & 73.85\% & 71 & -  & 75.35\% \cr
       
    
              \cmidrule(lr){1-2} \cmidrule(lr){3-5} \cmidrule(lr){6-8} \cmidrule(lr){9-11}
                \morecmidrules\cmidrule(lr){1-2} \cmidrule(lr){3-5} \cmidrule(lr){6-8} \cmidrule(lr){9-11}

                \multirow{4}{*}{\textbf{DCFL}}
                &FedAvg& 3  & 5  & 92.29 \% & 1  & 7  & 87.73\% & 17 & 25 & 85.90\% \cr 
       
                &FedProx&4  & 5  & 91.92 \% & 1  & 9  & 87.70\% & 8  & 19 & 85.96\% \cr
       
                &FedNova& 3  & 5  & 91.80 \% & 1  & 20 & 86.31\% & 25 & 31 & 84.90\% \cr  
       
                &FedDisco & 3  & 5  & 92.36 \% & 1  & 7  & 87.64\% & 16 & 25 & 86.37\%  \cr
                    
                    \bottomrule
        \end{tabular}  
                
    	\end{threeparttable} 
        }
        
\end{table*}  

\begin{figure*}[htbp]
    \centering
    \subfigure[Test accuracy under Dir$_{20}$(0.5)] {
        \begin{minipage}[b]{0.25\linewidth}
            \centering
            \includegraphics[scale=0.2]{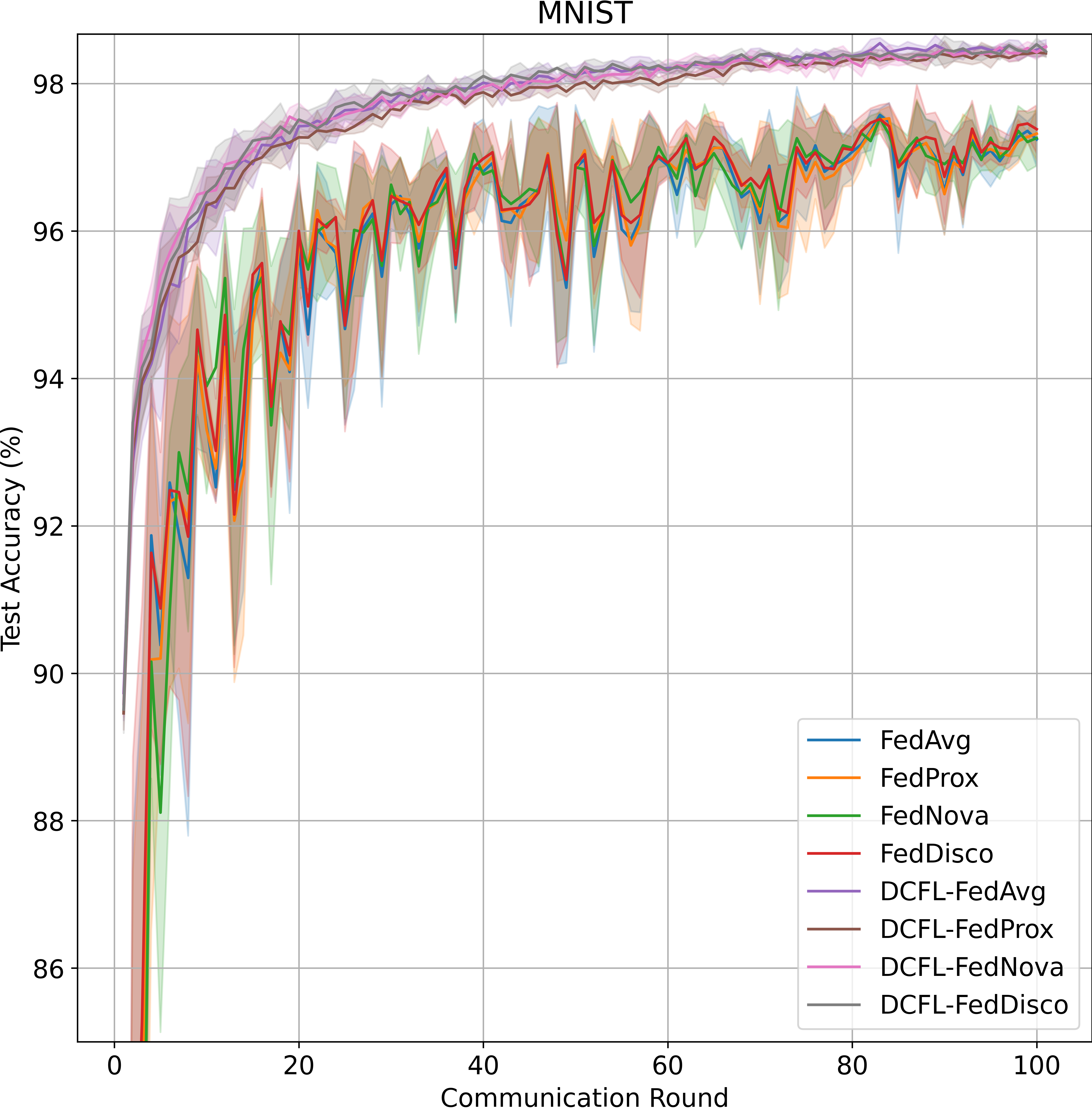}
        \end{minipage}
        \begin{minipage}[b]{0.25\linewidth}
            \centering
            \includegraphics[scale=0.2]{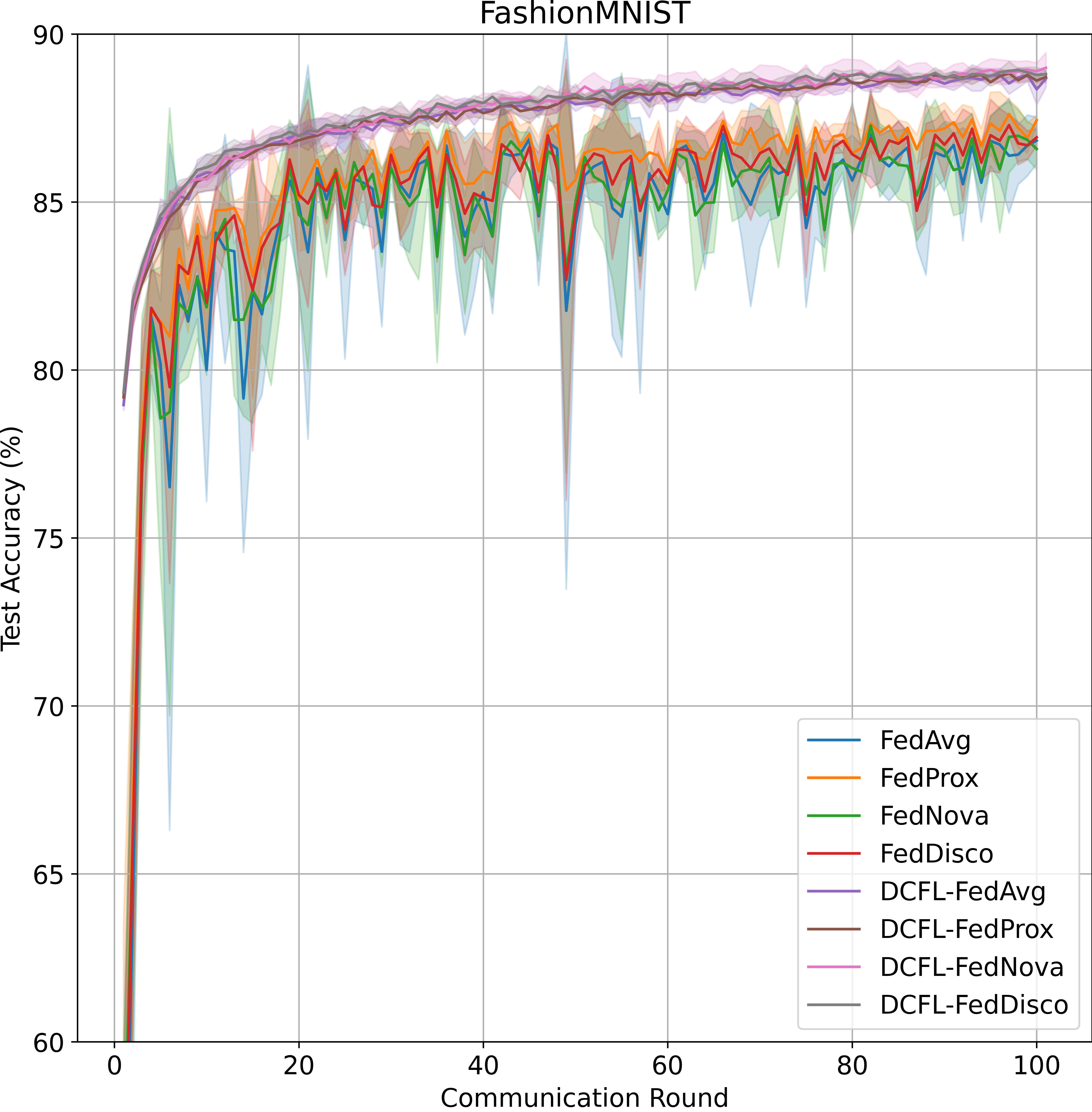}
        \end{minipage}
         \begin{minipage}[b]{0.25\linewidth}
            \centering
            \includegraphics[scale=0.2]{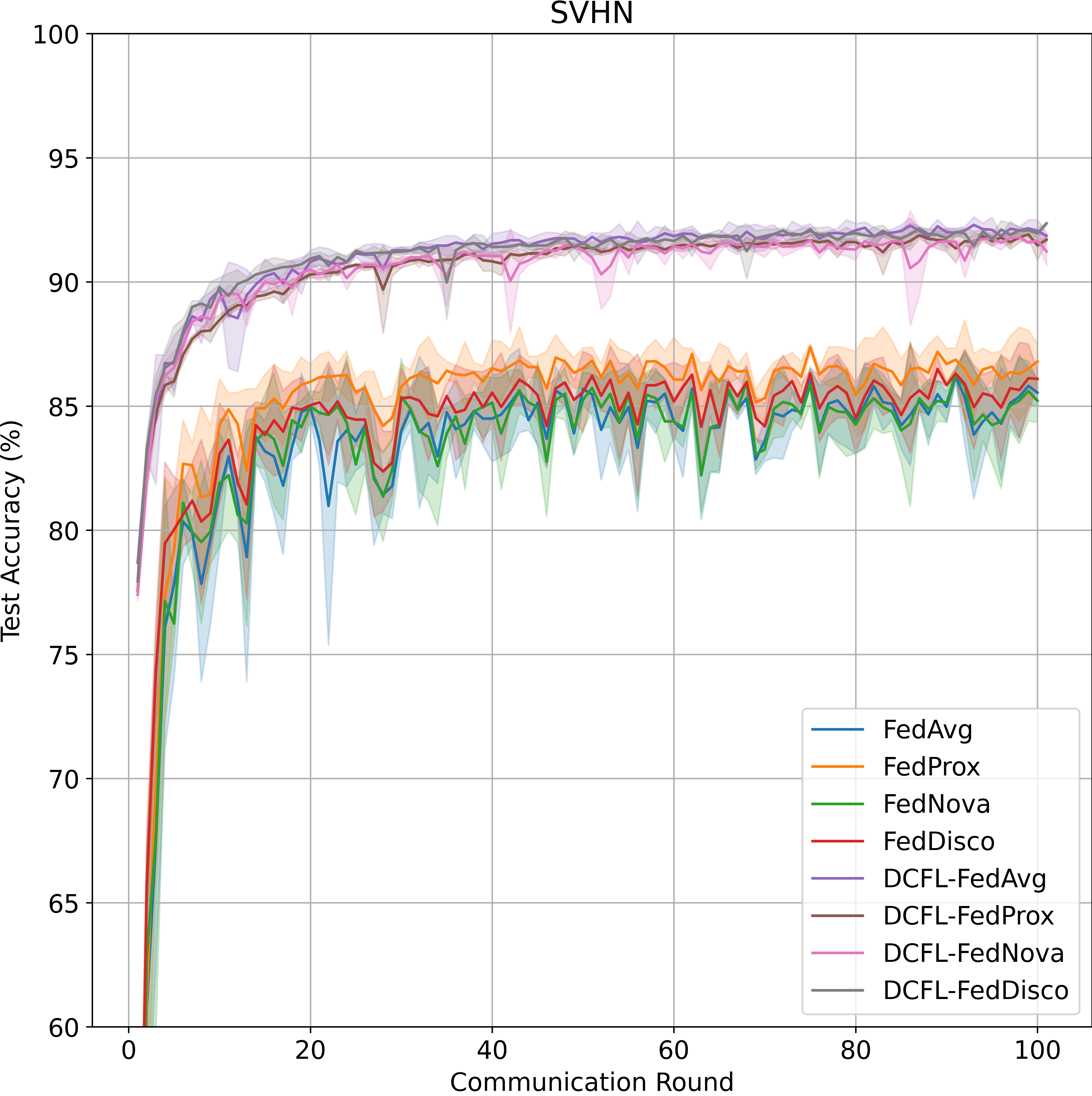}
        \end{minipage}
        \begin{minipage}[b]{0.25\linewidth}
            \centering
            \includegraphics[scale=0.2]{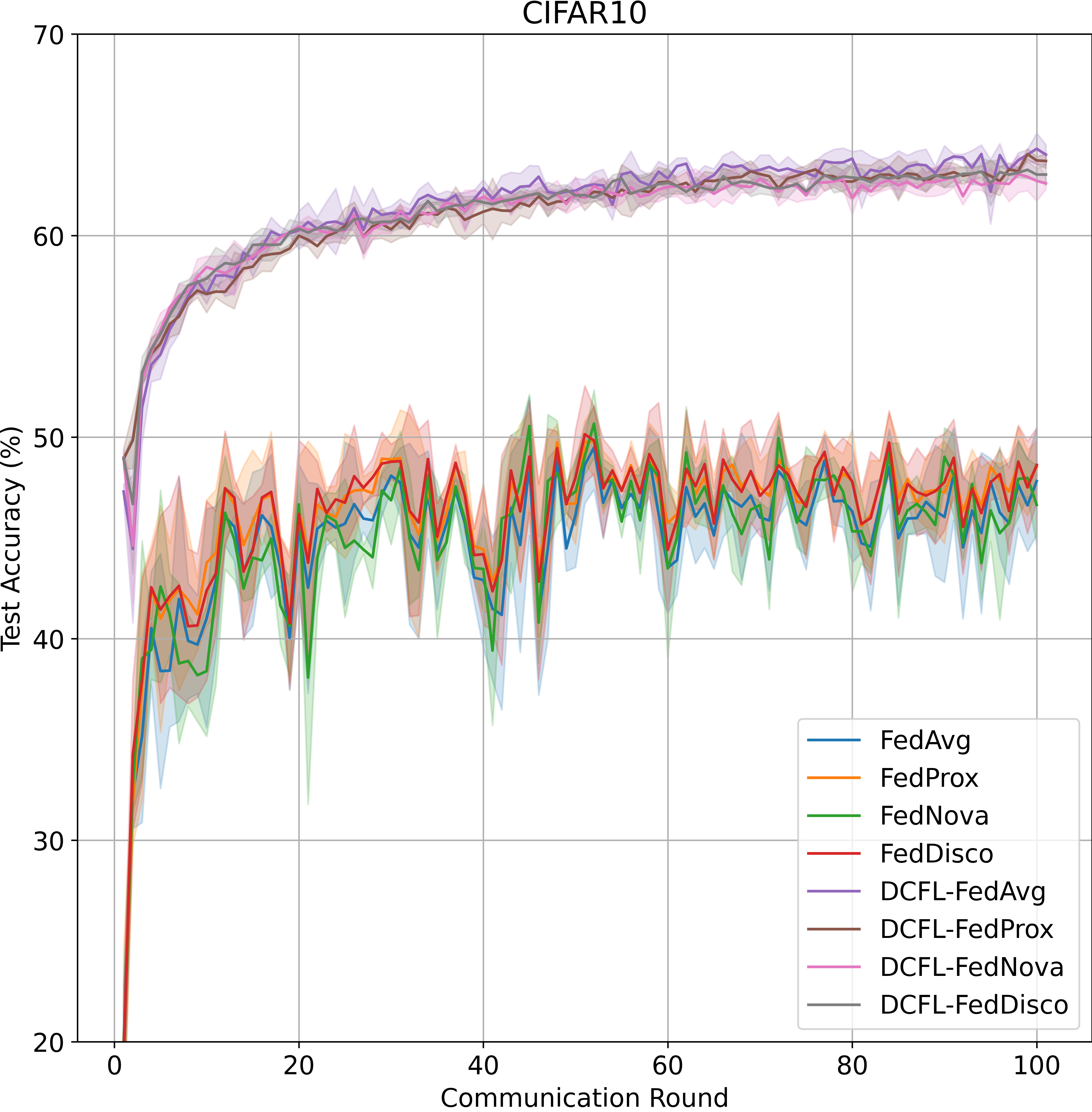}
        \end{minipage}
    }
    \subfigure[Test accuracy under Dir$_{20}$(0.1)]{
     \begin{minipage}[b]{0.25\linewidth}
            \centering
            \includegraphics[scale=0.2]{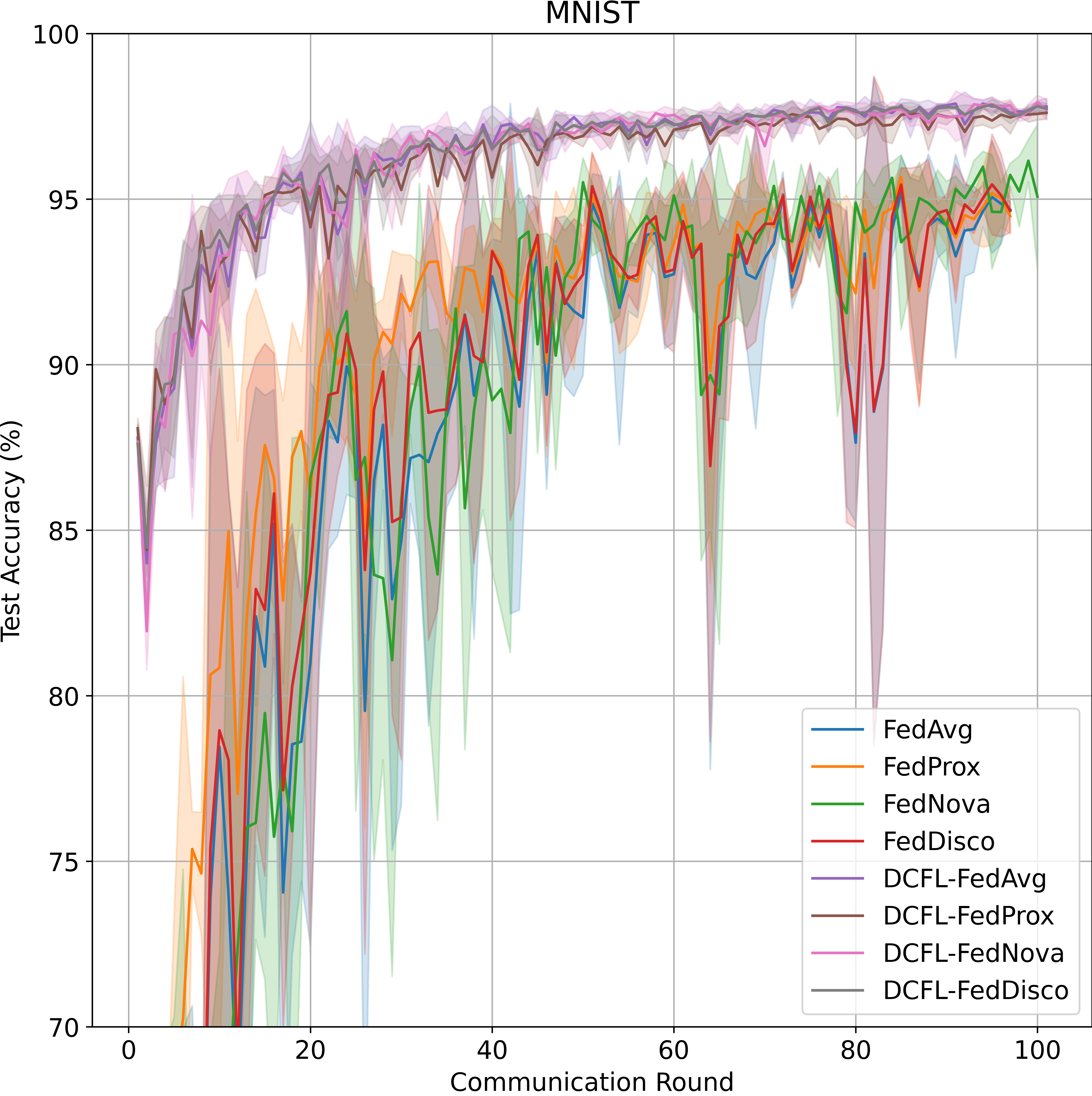}
        \end{minipage}
        \begin{minipage}[b]{0.25\linewidth}
            \centering
            \includegraphics[scale=0.2]{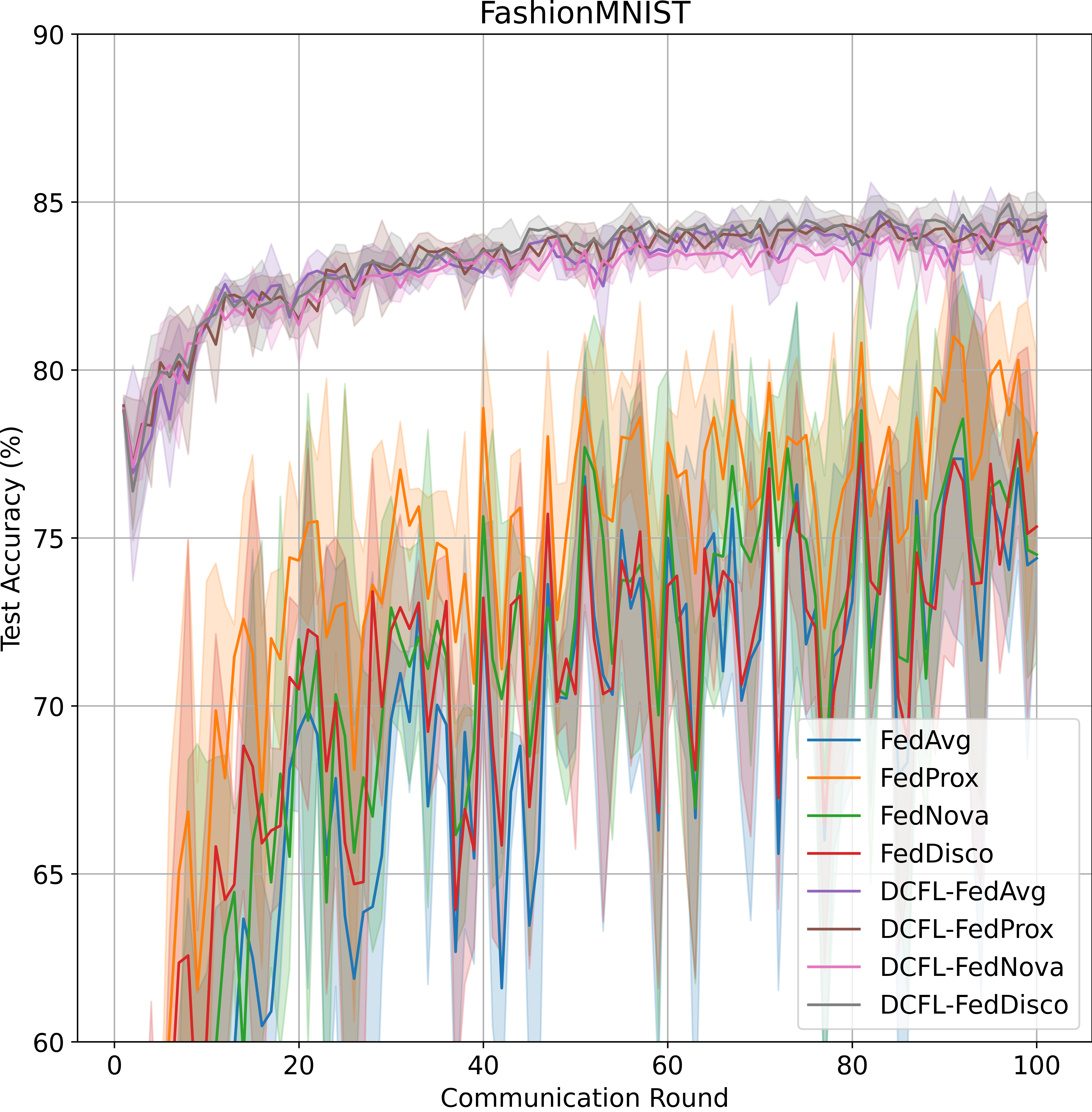}
        \end{minipage}
         \begin{minipage}[b]{0.25\linewidth}
            \centering
            \includegraphics[scale=0.2]{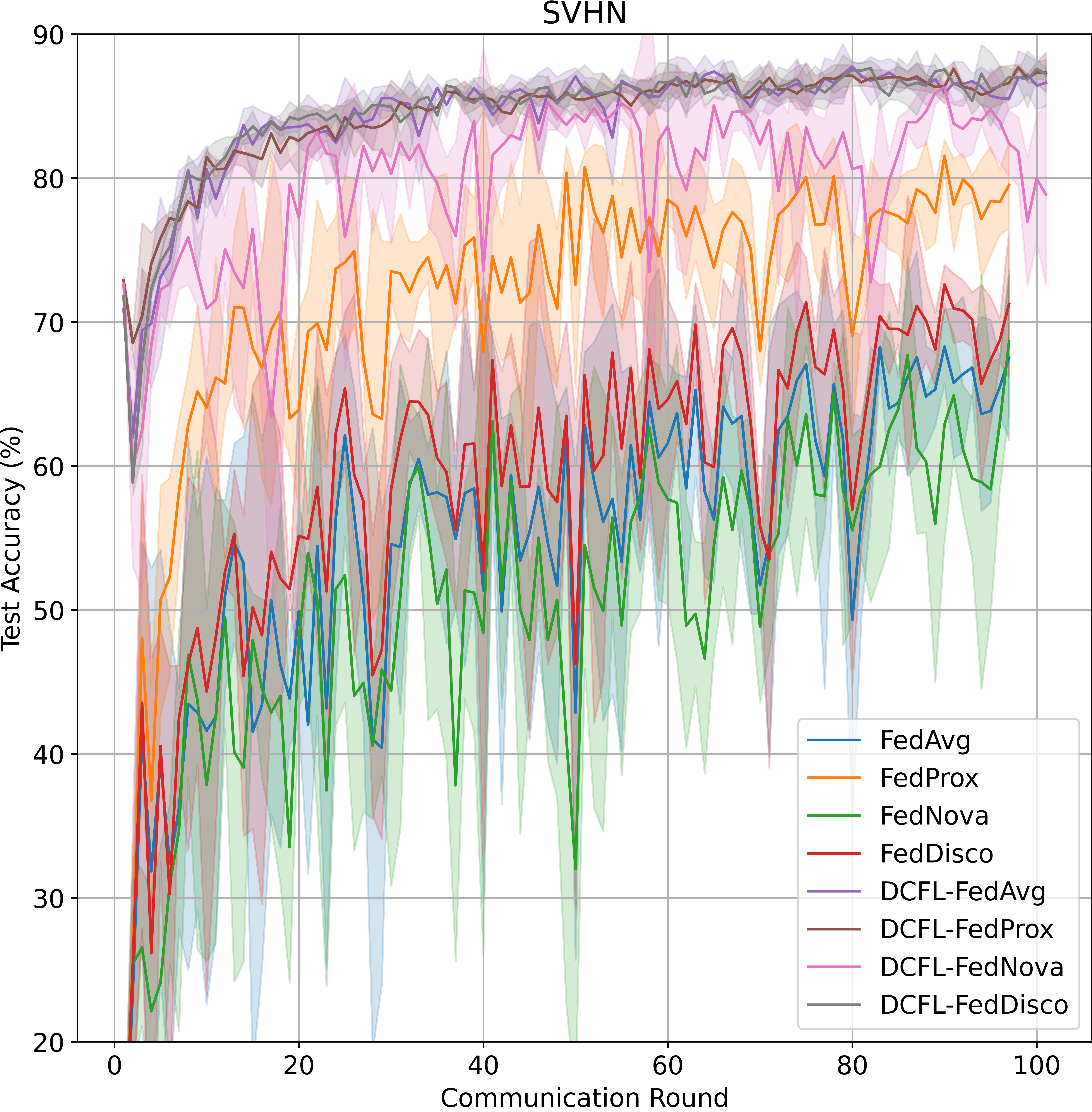}
        \end{minipage}
        \begin{minipage}[b]{0.25\linewidth}
            \centering
            \includegraphics[scale=0.2]{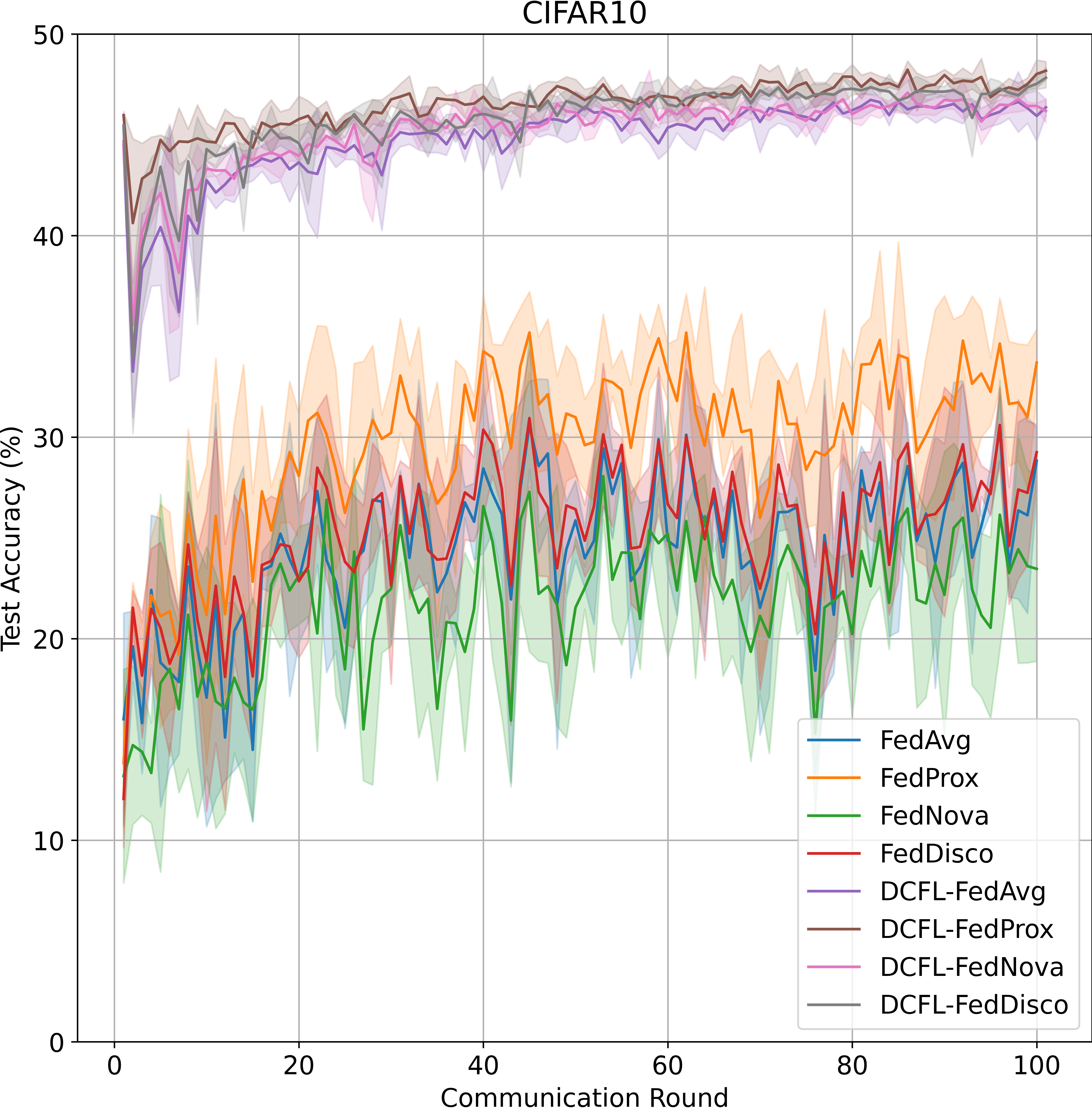}
        \end{minipage}
    }
    \subfigure[Test accuracy under C$_{k}$=2]{
     \begin{minipage}[b]{0.25\linewidth}
            \centering
            \includegraphics[scale=0.2]{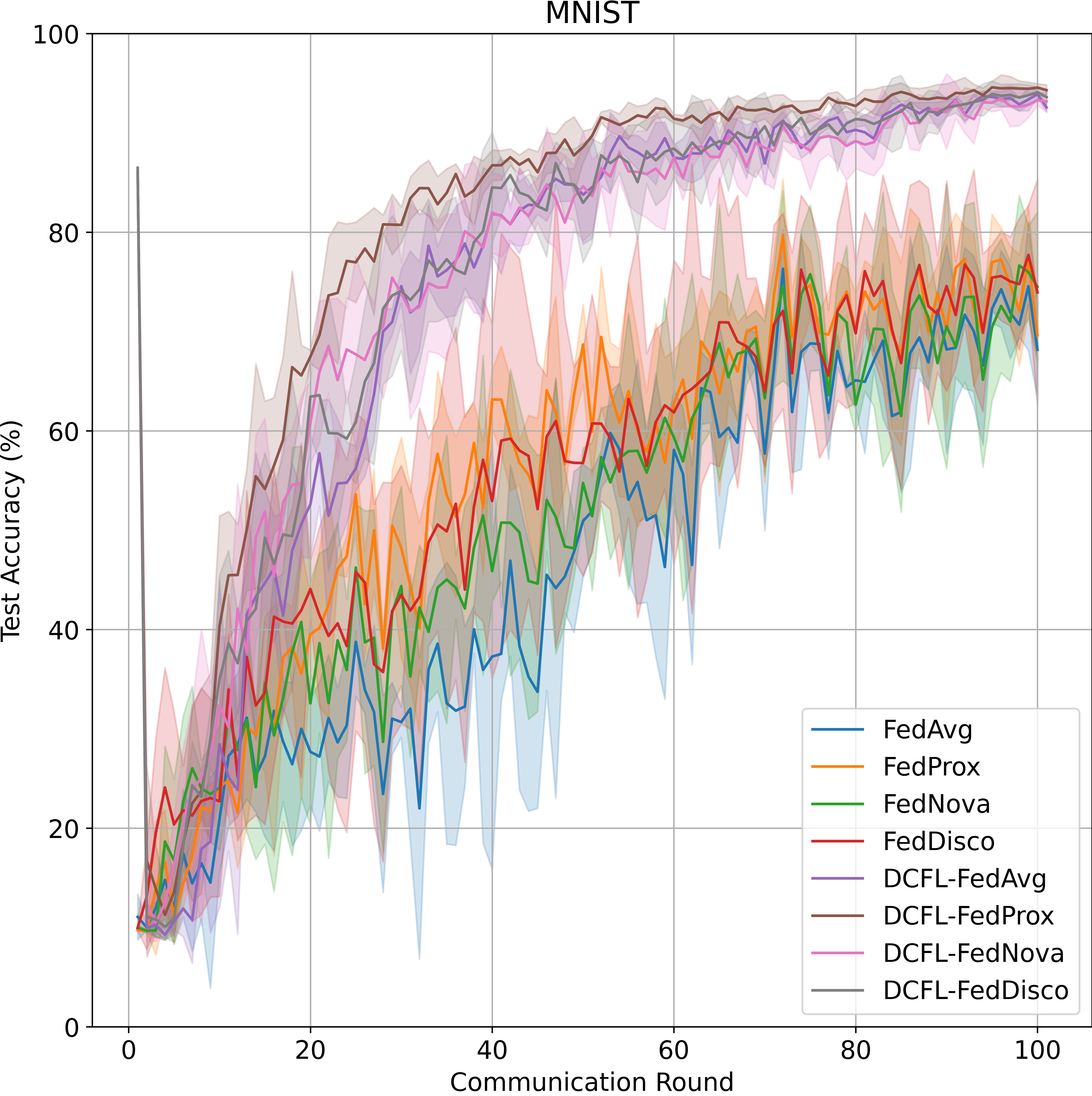}
        \end{minipage}
        \begin{minipage}[b]{0.25\linewidth}
            \centering
            \includegraphics[scale=0.2]{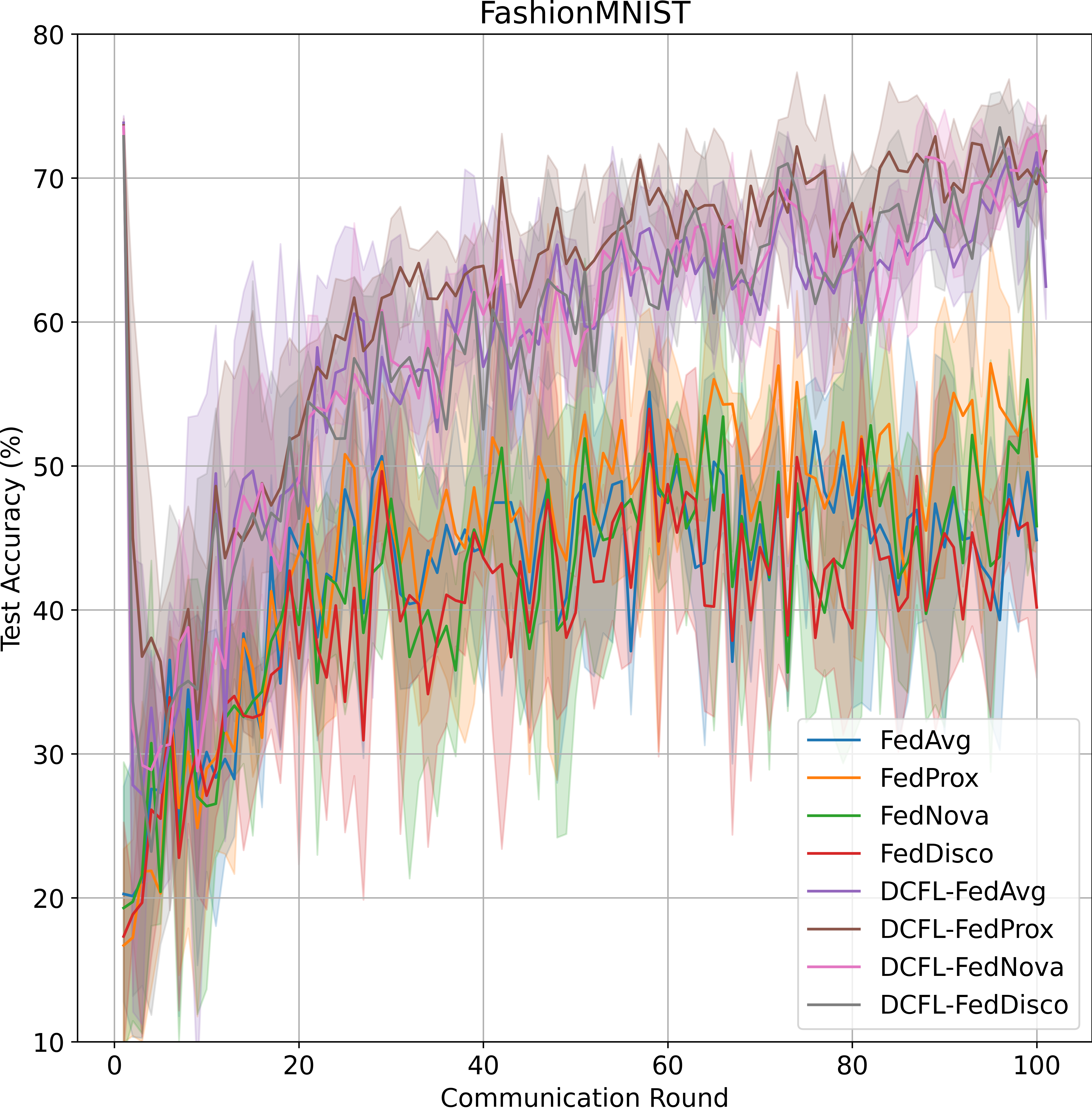}
        \end{minipage}
         \begin{minipage}[b]{0.25\linewidth}
            \centering
            \includegraphics[scale=0.2]{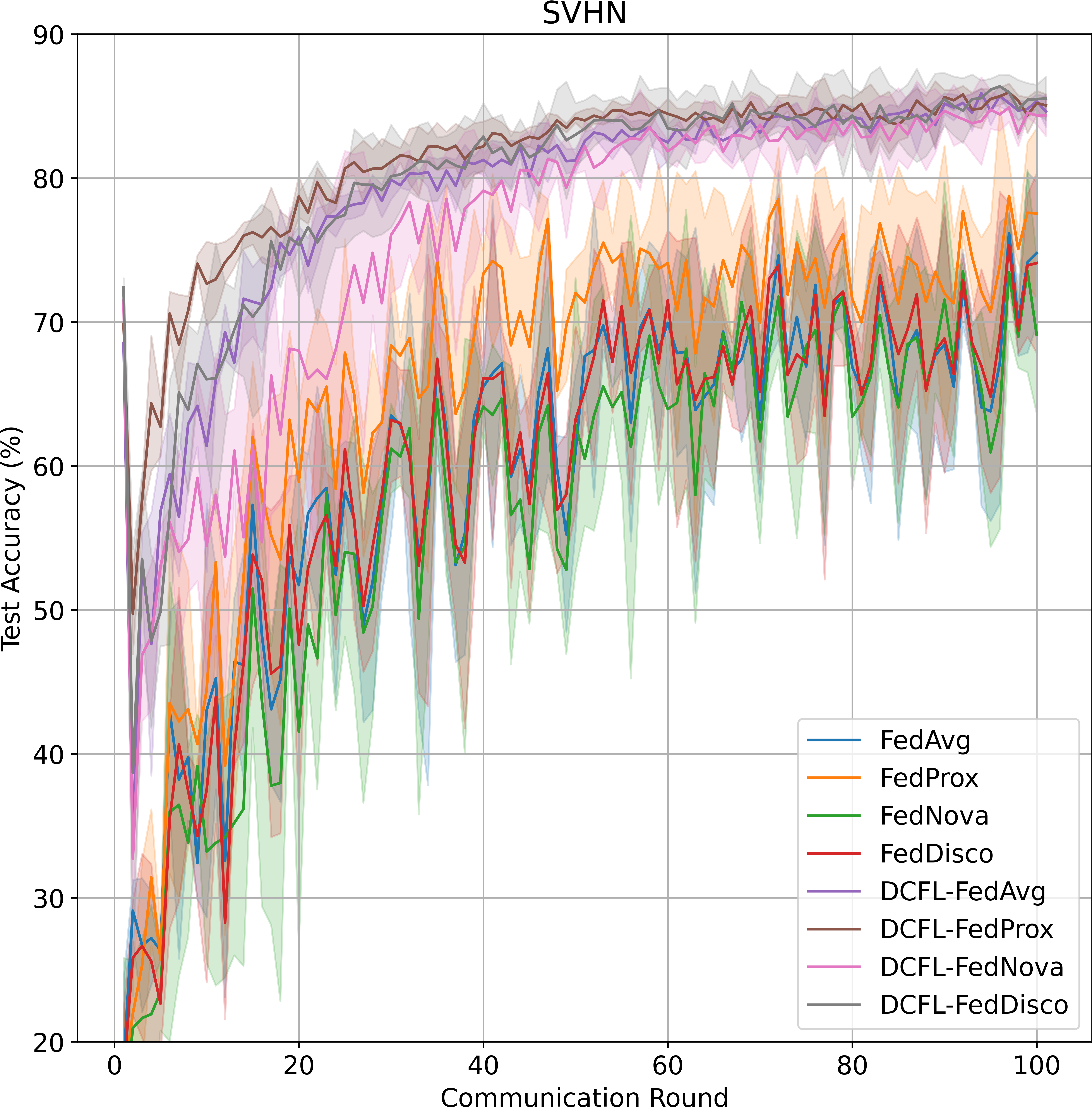}
        \end{minipage}
        \begin{minipage}[b]{0.25\linewidth}
            \centering
            \includegraphics[scale=0.2]{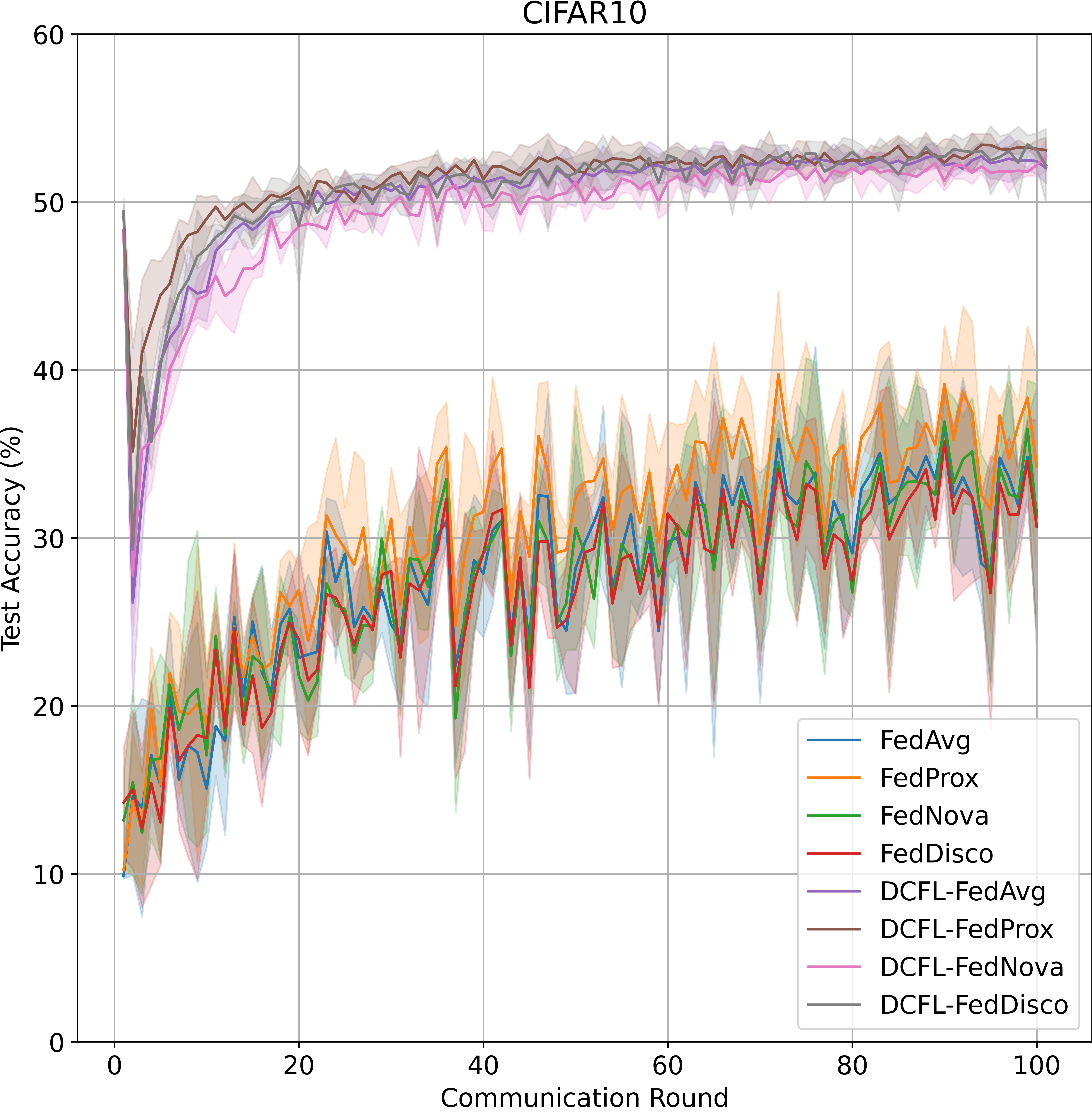}
        \end{minipage}
    }
    \caption{Test accuracy under different scenarios on four datasets}
    \label{fig:mnist_0.5dir}
\end{figure*}
\begin{table*}[ht]
\caption{Impact of each designed mechanism based on FedAvg Algorithm}
\label{tab::ablation_experiment}  
\centering
\begin{tabular}{ccccccc}
   
    \toprule
                 \multirow{2}{*}{Dataset}& \multicolumn{2}{c}{cka-guided client selection} & \multicolumn{2}{c}{Differentiable Siamese Augmentation} & \multicolumn{2}{c}{Fine-tuning} \\
                                    & w/o                     & w               & w/o                     & w & w/o                     & w                      \\
                        SVHN        & 70.58±2.06\%            & 76.87±1.32\%      & 70.58±2.06\%            & 74.08±3.12\%  
                        & 86.77±0.84\%            & 87.72±0.45\%                  \\
                       CIFAR10      & 29.21±1.61\%            & 33.25±1.56\%    & 29.21±1.61\%            & 35.08±2.18\%  
                       & 45.58±1.06\%            & 46.63±0.46\%                     \\
                       MNIST        & 95.33±0.57\%            & 95.74±0.43\%     & 95.33±0.57\%           & 96.43±0.62\%  
                       &  95.89±1.51\%            &  97.82±0.11\%              \\
                       FashionMNIST & 77.37±1.80\%           &  77.96±1.43\%      &  77.37±1.80\%            & 77.78±1.47\%  
                       & 78.50±1.86\%            & 84.44±0.28\%                \\
                        \bottomrule
\end{tabular}
\end{table*}

\subsection{Experimental setup}
\textbf{Datasets}. In this paper, we evaluate the image classification performance of the final model which is obtained through the whole process of DCFL. So we conduct experiments on four datasets including MNIST, Fashion MNIST~\cite{xiao2017fashion}, SVHN~\cite{netzer2011reading} and CIFAR10.
MNIST and Fashion MNIST consist of 28×28 gray-scale training images of 10 classes. SVHN and CIFAR10 contain 30k and 50k 32×32 training images from 10 categories respectively. We adopt the most representative and model-disturbing approach to divide the training data for clients - data partitioning with Dirichlet distribution Dir$_{k}$(a) (aka the label distribution skew~\cite{yang2019federated}). In this data split method, K is the number of clients and alpha determines the Non-IID level. A smaller value of alpha means a more unbalanced data distribution. We take two different scenarios of data distribution into consideration, including Dir$_{20}$(0.1), Dir$_{20}$(0.5). Besides, we also adopt quantity-based label imbalance data partition method (aka pathological Non-IID), like ~\cite{mcmahan2017communication}~\cite{li2022federated}, to extend a highly extreme scenario, where each client only has data samples with two label in our setting.

\textbf{Experimental Settings}. 
For CIFAR-10 and SVHN, we adopt a convolutional neural network (CNN), 
which is the typical deep neural network and commonly applied in FL. 
Specifically, we set two 5 x 5 convolution layers followed by max-pooling layers and two fully connected layers with ReLU activation, same to the structure mentioned by~\cite{mcmahan2017communication}~\cite{li2020federated}~\cite{wang2020tackling}~\cite{karimireddy2020scaffold}. For MNIST and Fashion MNIST, we adopt a simple multi-layer perceptron network.

\textbf{Hyper-parameters}. Like traditional FL training, DCFL also involves tuning a set of hyperparameters. 
Our method needs to tune a few additional hyper-parameters compare to traditional one, 
i.e. learning rate $\eta_{s}$ for the synthetic images, 
learning rate $\eta_{c}$ and local update epochs $E_{f}$ for the fine-tuning when using condensate data, 
filter ratio $r$ for condense data filtration,
exploration ratio $\epsilon$ for exploration-exploitation mechanism.
All experiments are run for three times with different random seeds with one NVIDIA 3080 GPU and
the average performance is reported in the paper

\subsection{Performance Comparison}
We first evaluate the performance of the proposed DCFL in improving the test accuracy, by comparing it with four baseline algorithms- FedAvg, FedProx, FedNova, and FedDisco.
According to Table \ref{tab::performance_comparison} , We can see that the accuracy of the server model trained by DCFL is generally higher than traditional FL methods in three settings. Compared with Dir$_{20}$(0.1), Dir$_{20}$(0.5), the data distribution of clients who obey pathological Non-IID is more extreme and heterogeneous.
In this context, the performance improved by the proposed DCFL is more significant than the other two scenarios. From table \ref{tab::performance_comparison}, we can see among these three schemes, the Federated Learning optimization method-FedAvg performs the worst, FedDisco performs relative well in total.
Compared with FedDisco, the proposed DCFL improves more than 1.01\%$\sim$16.95\%, 1.56\%$\sim$18.65\%, 5.49\%$\sim$14.14\%, and 13.24\%$\sim$18.34\% accuracy based on the four datasets, respectively.

\subsection{Communication Comparison}
Then, we evaluate the performance of the proposed DCFL in communication cost.
From table \ref{tab::communication_comparison}, We can see that the communication round of the server model cost to achieve specific accuracy by DCFL is generally lower than traditional FL methods in three settings.
Besides, we can see among these three schemes, the Federated Learning optimization method-FedAvg performs the worst, and FedProx performs relatively well in total.
Compared with FedProx, the proposed DCFL decreased more than 57, 82, and 53 rounds based on the SVHN dataset.

From another perspective, The communication volume in which the explicit message is uploaded to the server or downloaded from the server is relatively smaller in DCFL. This is especially obvious when training large neural network models, where the size of neural network parameters (or the gradient) is much larger than the size of transport condensed data. Take SVHN as an example, when the distribution of training data is Dir$_{20}$(0.1), the average number of classes per client (cpc) is 4.6. In our setting, we adopt the number of images per class of 10 for obtaining condensed data, so the total number of condensed data's float parameters uploaded to the server can be generally considered: the number of clients $\times$ cpc $\times$ ipc $\times$ image size = 20 $\times$ 4.6 $\times$ 10 $\times$ 3 $\times$ 32 $\times$ 32 $\approx$ 5.2 × 10$^{6}$, while the size of each complementary group is set to same to participate clients per round, which is the number of clients $\times$ sample ratio. So the data size downloaded from the server can be roughly considered: the number of clients $\times$ (the number of clients $\times$ sample ratio - 1) $\times$ cpc $\times$ ipc $\times$ image size = 20 $\times$ (20 $\times$ 0.25 - 1) $\times$ 4.6 $\times$ 10 $\times$ 3 $\times$ 32 $\times$ 32 $\approx$ 20.8 × 10$^{6}$.So the extra float parameters in DCFL is 2.6 $\times$ 10$^{7}$. For those iterative model averaging model methods, the number of float parameters is equal to the product of weight size and the number of participating clients, which is the model parameters of model $\times$ (the number of clients $\times$ sample ratio) $\times$ 2 $\approx$ 3.2 × 10$^{6}$ for ConvNet. Although the volume of condensed data seems large, DCFL can reduce the total communication round around 15x$\sim$20x, so in total we can reduce the communication volume more than 301 $\sim$ 557 MB.
 
\subsection{Component-wise Analysis}
Next, we implement three breakdown versions of DCFL to evaluate and understand the effectiveness of each of the key components incorporated in DCFL.

\textbf{Effects of CKA-guided client selection}. The CKA-guided client selection strategy of DCFL guarantees the finer-grained client selection by considering the data distribution similarity between clients. Without the auxiliary data transportation and other tricks, Table \ref{tab::ablation_experiment} shows the comparison result of the CKA-guided client selection strategy and the traditional random selection strategy. From this, we can see the improvements in final accuracy for all datasets, with around 0.41\% $\sim$ 6.29\%, respectively.

\textbf{Effects of Differentiable Siamese Augmentation}. DSA is a family of image transformations that preserves the semantics of the input such as cropping, color jittering, and flipping that are parameterized with model $w$ for the synthetic and real training sets respectively. By using DSA in DCFL, this strategy enables correspondence between the two sets and provides a more effective way of exploiting the information in the real training images and condensed images. Table \ref{tab::ablation_experiment} shows the comparison between with the DSA strategy and without the DSA strategy. From this, we can see the improvements in final accuracy for all datasets, with around 0.41\% $\sim$ 6.29\%, respectively.

\textbf{Effects of fine-tuning}. DCFL utilizes condensed data effectively while stabilizing and improving the model performance via applying fine-tuning. In DCFL, when participating clients receive complementary condensed data, they use the local dataset to pre-train, and then use auxiliary data to fine-tune, compare with other method, like mingling condensed data and local data as training dataset. Table \ref{tab::ablation_experiment} shows that the improvements in accuracy can reach 0.95\% $\sim$ 5.94\% for DCFL with fine-tuning.

\section{Relative Work}
\begin{figure}[ht]
    \centering
    \includegraphics[width=1\linewidth]{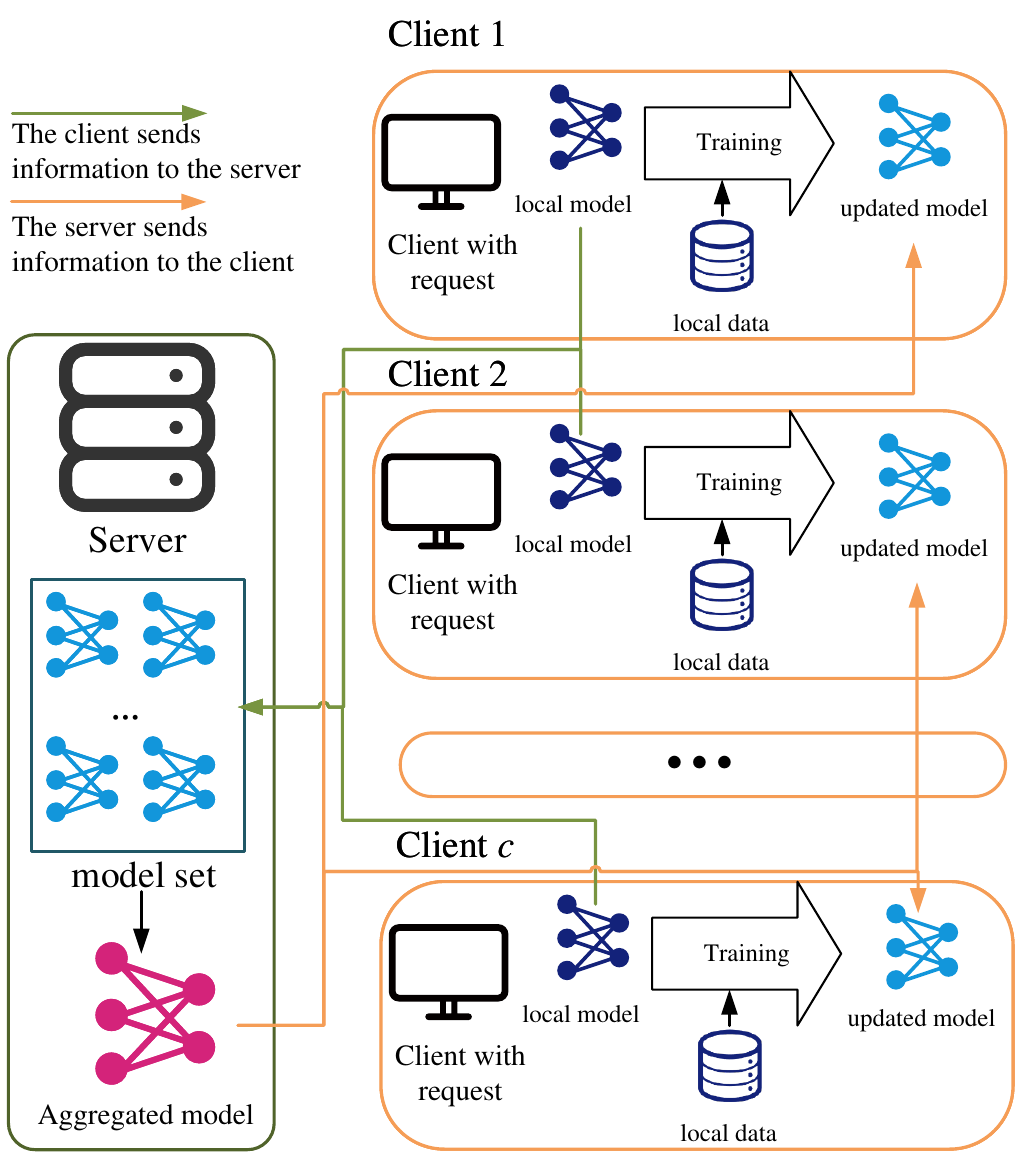}
    \caption{General flow of federated learning.}
    \label{fig::flowchart_fl}
\end{figure}
\subsection{Federated Learning}
Federated learning, as a variation of distributed optimization, has attracted more and more attention and application nowadays in research and industry areas~\cite{kairouz2021advances}. According to Figure \ref{fig::flowchart_fl}, the traditional execution flow of FL can be deconstructed in a few necessary steps:
\begin{enumerate}
    \item Initialization. To disseminate model parameters to participants, the parameter server generates a random or pre-trained global model.
    \item Participant Selection. The server selects a fraction number of participants to participate in the current round training process.
    \item Download. The selected participants download the latest model from the server side.
    \item Local Update and Upload. Selected participants locally train the transmitted model by using their private data samples and upload the trained model parameters to the server.
    \item Aggregation. After receiving the local models from selected participants, the server aggregates them and computes a new global model.
\end{enumerate}

According to the distribution characteristics of clients’ private datasets, the category of FL can be divided into horizontal FL, vertical FL, and Federated transfer learning~\cite{yang2019federated}. In this paper, we mainly focus on the horizontal FL setting.

\subsection{Dataset Condensation}
Dataset condensation\footnote{It is same to dataset distillation.}
can be used to construct a smaller but informative synthetic dataset from the original large training dataset, which condensed data is different from original training data and can acquire a comparable generalization performance with less training cost~\cite{wang2018dataset}.
Based on the objectives applied to mimic target data or to find a proxy model that learns synthetic datasets by optimizing their features and corresponding decoders, dataset condensation methods can be divided into a Meta-Learning Framework, Data Matching Framework, and Factorized Dataset Distillation~\cite{lei2023comprehensive}.
Some works, like~\cite{nguyen2020dataset} use kernel to get synthetic data, while effective but time-consuming;
Some works, like~\cite{cazenavette2022dataset} directly matched the long-range trajectory between the target dataset and the synthetic dataset instead of single gradient matching, so that the computational overhead of training and storing expert trajectories is quite high.
Due to the comprehensive consideration of computing cost, memory usage, and the specific features of FL, we use DC~\cite{zhao2020dataset}, DM~\cite{zhao2023dataset}, DSA~\cite{zhao2021dataset} as our dataset condensation methods in DCFL.

\subsection{Centered Kernel Alignment}
To better understand and characterize the neural network representations learned from data, researchers from Google proposed the novelty and insightful method named Centered Kernel Alignment~\cite{kornblith2019similarity}. CKA provides an effective way to measure similarities between deep neural network representations, which takes the complex interaction between the training dynamics and structured data into account. Few pieces of literature that apply CKA to Federated learning, but rarely of they further consider the use of the derived CKA indexes to reflect the complementary of privately owned datasets between different clients. For example, Mi Luo et al.~\cite{luo2021no} used the CKA to measure the similarity between the representations from the same layer of different clients’ local models and found there exists a greater bias in the classifier than other layers, they didn’t use that information to further deduce the peer-to-peer dataset complementary relationships, while to post-calibrate the classifier after federated training. In this paper, we use CKA from a novel perspective to guide client selection and condensed data utilization.

\section{Conclusion}
In this paper, we propose a novel implementation framework of federated learning - DCFL to achieve faster convergence, stabilize the model training process, and better model performance. By using the CKA-based client complementarity method to guide group division clients and then condensed data-assisted client model training with Non-IID awareness, we have reduced the communication rounds to reach convergence. However, there is still a need for relatively costly computation time to obtain condensed data locally, and can’t apply complex datasets, like CIFAR-100, Tiny ImageNet and ImageNet, in DCFL due to the limitations of current data condensation methods. How to reduce the computation time to get the synthetic set and apply the framework to more complex datasets can be potential future directions.

\section*{Acknowledgment}

The preferred spelling of the word ``acknowledgment'' in America is without 
an ``e'' after the ``g''. Avoid the stilted expression ``one of us (R. B. 
G.) thanks $\ldots$''. Instead, try ``R. B. G. thanks$\ldots$''. Put sponsor 
acknowledgments in the unnumbered footnote on the first page.

\bibliography{ref}
\bibliographystyle{IEEEtran}

\end{document}